\documentclass[10pt,twocolumn,letterpaper]{article}
\ifdefined\pdfobjcompresslevel
\fi
\usepackage[pagenumbers]{cvpr} %

\DeclareMathOperator*{\argmax}{arg\,max}

\definecolor{cvprblue}{rgb}{0.21,0.49,0.74}
\usepackage[pagebackref,breaklinks,colorlinks,allcolors=cvprblue]{hyperref}

\def\confName{CVPR}
\def\confYear{2027}

\usepackage[ruled,vlined]{algorithm2e}
\usepackage{booktabs}       %
\usepackage{siunitx}        %
\usepackage{multirow}       %
\usepackage{amsmath}        %
\usepackage{array} %
\usepackage{graphicx} %
\usepackage{amssymb}        %
\usepackage{bbding}
\usepackage{pifont}
\usepackage{wasysym}
\usepackage{utfsym}
\usepackage{fontawesome}
\usepackage{colortbl}
\usepackage{tikz}
\usepackage{balance}
\usepackage{flushend}
\definecolor{red}{HTML}{D2042D}
\newcommand*{\circled}[1]{\lower.7ex\hbox{\tikz\draw (0pt, 0pt)%
    circle (.38em) node {\makebox[1em][c]{\small #1}};}}

\title{\textbf{\mbox{\textcolor[HTML]{FF3030}{A}\textcolor[HTML]{F39C12}{c}\textcolor[HTML]{E4B900}{t}\textcolor[HTML]{00A651}{W}\textcolor[HTML]{0072BC}{o}\textcolor[HTML]{00A5CF}{r}\textcolor[HTML]{A142AF}{ld-JEPA}}}: Action-Conditioned JEPA World Models Meet\\
Latent Planning for End-to-End Autonomous Driving}

\author{
Ziying Song$^{1,\dagger}$,
Shengkai Zhang$^{2,\dagger}$,
Caiyan Jia$^{2}$,
Lei Yang$^{1}$,
Lin Liu$^{3}$,\\
Yuchen Liu$^{4}$,
Ziyang Liu$^{5}$,
Zirui Li$^{1}$,
Gang Wang$^{6}$,
Peiliang Wu$^{7}$,\\
Li Wang$^{8}$,
Yadan Luo$^{9}$,
Chen Lv$^{1,*}$\vspace{1ex}\\
$^{1}$Nanyang Technological University\quad
$^{2}$Beijing Jiaotong University\\
$^{3}$Dalian University of Technology\quad
$^{4}$North University of China\\
$^{5}$Tsinghua University\quad
$^{6}$NingboTech University\quad
$^{7}$Yan Shan University\\
$^{8}$Beijing Institute of Technology\quad
$^{9}$The University of Queensland\\[0.8ex]
{\normalsize\mbox{\faGithub~Code:
\href{https://github.com/adept-thu/ActWorld-JEPA}{\texttt{ActWorld-JEPA}}\qquad
\textcolor{cvprblue}{\faGlobe}~Project:
\href{https://adept-thu.github.io/ActWorld-JEPA/}{\texttt{adept-thu.github.io/ActWorld-JEPA}}}}
}
\hypersetup{pdfauthor={Ziying Song, Shengkai Zhang, Caiyan Jia, Lei Yang, Lin Liu, Yuchen Liu, Ziyang Liu, Zirui Li, Gang Wang, Peiliang Wu, Li Wang, Yadan Luo, Chen Lv},pdftitle={ActWorld-JEPA: Action-Conditioned JEPA World Models Meet Latent Planning for End-to-End Autonomous Driving}}

\begin{document}
\maketitle
\begingroup
\renewcommand{\thefootnote}{}
\footnotetext{$^{\dagger}$ Co-first authors. $^{*}$ Corresponding author.}
\endgroup
\begin{abstract}
Joint-Embedding Predictive Architectures (JEPAs) enable future prediction
in latent space without pixel reconstruction.
However, latent predictive consistency alone does not establish
the relative driving quality of candidate plans.
We present \textbf{ActWorld-JEPA}, a joint-embedding formulation for
candidate-based driving. Where \textbf{Plan-Conditioned JEPA (PC-JEPA)} uses global and step-wise
trajectory intentions to guide recurrent latent prediction, then combines
the predicted future with the current scene through gated cross-attention;
\textbf{Relational Residual Reranking ($R^3$)} models agreement and
discrepancy between plan and future representations under planning
supervision, yielding bounded, gated residual corrections to the base
planning scores. This formulation connects latent prediction to candidate
selection using only observed context and proposed trajectories at
inference. Benchmark-specific implementations report \textbf{94.1 PDMS}
on NAVSIM v1 navtest, with the final selector calibrated using aggregate
NAVTEST feedback, and \textbf{91.6} and \textbf{42.7 EPDMS} on NAVSIM v2
navtest and navhard, respectively. The nuScenes evaluation reports a mean
$L_2$ error of \textbf{1.20~m} and a predicted collision rate of
\textbf{0.76\%} over 1--6~s. This work provides an effective path from
JEPA's predictive representations to driving decisions informed by
anticipated consequences. 
\end{abstract}

\begin{figure*}[t]
    \centering
    \includegraphics[width=\textwidth]{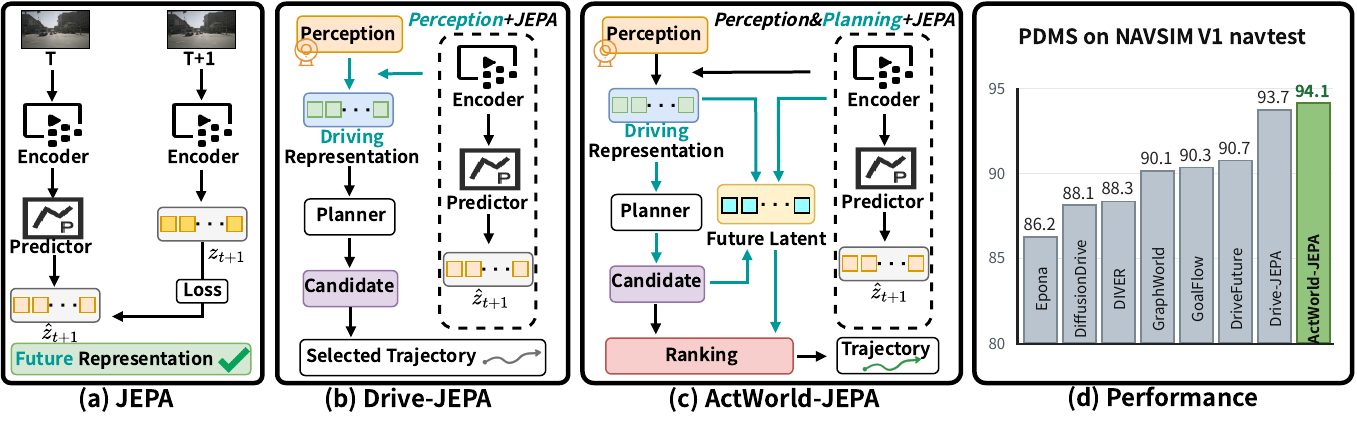}
    \vspace{-1ex}
    \caption{
\textbf{From predictive representations to candidate-conditioned planning.}
\textbf{(a) JEPA-style representation learning} predicts target features
from observed context, as in V-JEPA~\cite{bardes2024vjepa}.
\textbf{(b) Drive-JEPA}~\cite{wang2026drive} combines video-JEPA
representations with multimodal trajectory distillation and
momentum-aware selection.
\textbf{(c) ActWorld-JEPA} conditions latent prediction on candidate plans
and relates the predicted futures to trajectory ranking through PC-JEPA
and $R^3$. Both observed scene context and candidate intention contribute
to the decision. 
\textbf{(d) Planning performance.}
PDMS on NAVSIM v1 navtest~\cite{navsimv1}, comparing ActWorld-JEPA
with Epona~\cite{zhang2025epona}, DiffusionDrive~\cite{liao2024diffusiondrive},
DIVER~\cite{diver}, GraphWorld~\cite{song2026graphworld},
GoalFlow~\cite{xing2025goalflow}, DriveFuture~\cite{hong2026drivefuture},
and Drive-JEPA~\cite{wang2026drive}.
}
\label{fig:motivation}
\end{figure*}

\section{Introduction}
\label{introduction}
End-to-end autonomous driving brings scene understanding and motion
planning into a common framework that maps sensor observations to
driving trajectories~\cite{LiHongyange2etpamisurvey,uniad,jiang2023vad}.
Recent planners represent possible maneuvers through trajectory
proposals~\cite{li2024hydra,guo2026ipad} or generative
sampling~\cite{liao2024diffusiondrive,xing2025goalflow}. Choosing among
them requires weighing safety, comfort, and progress as traffic
evolves~\cite{navsimv1,navsimv2}. Consider a vehicle approaching a merge
as another enters from a ramp. Maintaining speed and slowing down may
both be feasible initially, yet the two maneuvers leave different gaps
for the merging vehicle a few seconds later.

World models offer a way to anticipate such interactions by predicting
how a scene may evolve under a proposed
action~\cite{feng2025survey_dwm_survey,guan2024world_dwm_survey}.
Future images depict changes in the scene from the ego vehicle's
viewpoint~\cite{wang2024drivedreamer,wang2024driving}, while occupancy
predictions describe how occupied regions change over
time~\cite{zheng2024occworld}. Models that predict compact latent
features can supply future scene information directly to a
planner~\cite{li2024enhancing_law,zheng2025world4drive,wote,song2026momworld}.
Conditioning these predictions on individual candidates allows a
planner to compare how different maneuvers might unfold from the same
initial scene.

Joint-Embedding Predictive Architectures (JEPAs) offer a latent-space
approach to world modeling by learning to predict target embeddings
from observed context~\cite{assran2023ijepa,bardes2024vjepa} (Figure~\ref{fig:motivation} (a)). This
objective supports self-supervised learning from video. With an
action-conditioned predictor, a planner can compare candidate futures
in feature space without decoding each rollout into
images~\cite{assran2025vjepa2,maes2026leworldmodel,zhou2025dinowm}.
Drive-JEPA uses V-JEPA to pretrain a visual encoder on driving videos
and transfers its features to a proposal-based planner~\cite{wang2026drive}.
The planner scores proposal features directly, without predicting a
future scene representation conditioned on each trajectory (Figure~\ref{fig:motivation} (b)). Auto-JEPA
uses predicted driving intent to retrieve trajectories for
scene-conditioned ranking~\cite{yang2026autojepa}. AD-E2E-JEPA compares
candidate rollouts with a goal image~\cite{zhu2026ade2ejepa}, and DA-WAM
estimates planning utility from each candidate's predicted
future~\cite{zhong2026dawam}. These approaches highlight a distinction
between forecasting an outcome and judging its value for driving.
At the merge, a predictor could accurately anticipate the gap left by
either maneuver. Choosing between them requires weighing that clearance
against the progress lost by slowing down. %
A promising solution is to use planning supervision
to learn how the relation between each candidate's intended motion and
its predicted future should refine the base planner's score.

Thus, we propose \textbf{ActWorld-JEPA}, a joint-embedding formulation that
connects candidate-conditioned future prediction with relational
trajectory ranking, as illustrated in Figure~\ref{fig:motivation} (c).
\textbf{Plan-Conditioned JEPA (PC-JEPA)} encodes
each candidate into global and step-wise intentions. The global
intention modulates the initial latent state, and the step-wise
intentions condition recurrent prediction. Gated cross-attention then
integrates the predicted rollout with the current scene to obtain a
candidate-specific, future-aware representation. In the proposed JEPA
objective, future observations supply stop-gradient target embeddings
for the corresponding observed trajectory. Inference uses only
available historical observations and candidate plans.
\textbf{Relational Residual Reranking ($R^3$)} models agreement and
discrepancy between the candidate intention and its future-aware
representation. A planning-supervised relation head produces a learned
gate and a bounded score residual. Together, these components connect
trajectory intention, latent future prediction, and score refinement,
allowing predictive information to influence candidate preferences.

We conduct benchmark-specific experiments on NAVSIM
v1/v2~\cite{navsimv1,navsimv2} and nuScenes~\cite{nuscenes}, covering
non-reactive pseudo-simulation and long-horizon open-loop planning.
The evaluated systems report 94.1 PDMS on NAVSIM v1 navtest and
91.6 and 42.7 EPDMS on NAVSIM v2 navtest and navhard, respectively.
On nuScenes, they achieve an average $L_2$ error of 1.20~m and a
predicted collision rate of 0.76\% across the 1--6~s horizons.
Our contributions are summarized as follows:
\begin{itemize}
    \item We propose ActWorld-JEPA, a joint-embedding predictive
    framework for end-to-end autonomous driving. It connects
    plan-conditioned latent future modeling with trajectory selection,
    enabling candidate decisions to incorporate predicted consequences.

    \item We introduce two complementary modules, PC-JEPA and $R^3$.
    PC-JEPA uses global and step-wise trajectory intentions to predict
    latent futures and integrate them with the current scene.
    $R^3$ models plan--future relations to refine candidate scores
    through bounded, gated residuals.

    \item We evaluate ActWorld-JEPA on NAVSIM v1/v2 and nuScenes,
reporting 94.1 PDMS on v1 navtest and 91.6/42.7 EPDMS on
v2 navtest/navhard (leaderboards:
\href{https://huggingface.co/spaces/AGC2024-P/e2e-driving-navtest}{v1 navtest},
\href{https://huggingface.co/spaces/AGC2025/e2e-driving-navhard}{v2 navhard}).
\end{itemize}

\begin{figure*}[t]
\centering
\includegraphics[width=\linewidth]{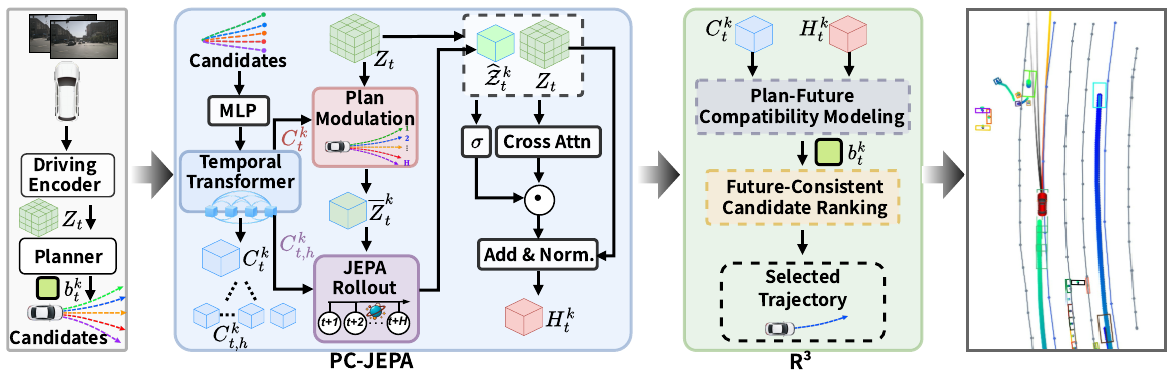}
\caption[ ]{
\textbf{The overall architecture of ActWorld-JEPA.}
The driving encoder extracts the current representation, and the planner
produces candidate trajectories. PC-JEPA encodes global and step-wise plan
conditions, initializes plan-modulated latents, and rolls out
candidate-specific futures. Cross-attention and gated residual interaction
fuse predicted future evidence with the current scene. $R^3$ combines plan
and fused-future features to rank the original candidates. The rightmost
trajectory panel illustrates future-guided selection.
}
\label{fig:main}
\end{figure*}

\begin{figure}[t]
\centering
\includegraphics[width=\linewidth]{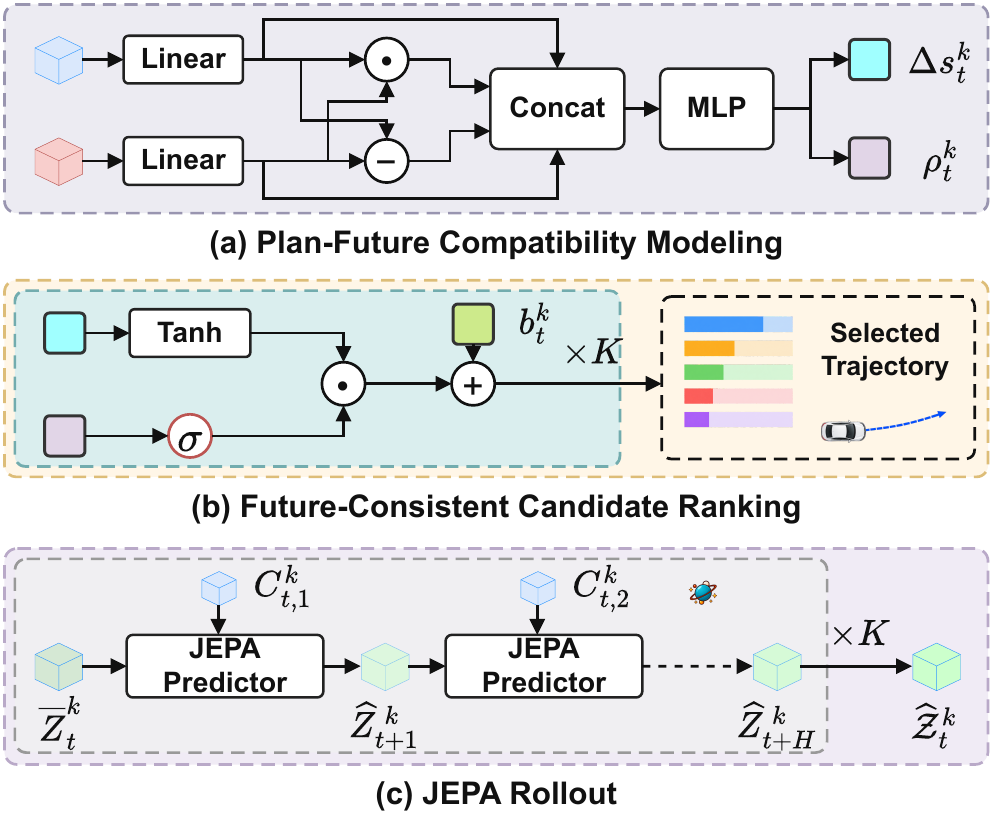}
\caption[ ]{
\textbf{Details of latent prediction and trajectory reranking.}
(a) Plan-future compatibility modeling combines projected features and
their agreement and discrepancy to predict a score correction $\Delta s_t^k$
and confidence logit $\rho_t^k$.
(b) Future-consistent candidate ranking gates the bounded correction,
adds the baseline score $b_t^k$, and selects an original trajectory.
(c) JEPA rollout predicts latent futures conditioned on step-wise
trajectory intentions over the planning horizon.
}
\label{fig:sub2}
\end{figure}

\section{Related Work}
\label{related_work}

\subsection{World Models for Autonomous Driving}

World models anticipate scene evolution to support prediction and planning
in autonomous driving. Visual generative approaches synthesize controllable
future observations through diffusion or autoregressive modeling for
action-conditioned planning~\cite{hu2023gaia1,wang2024drivedreamer,
hu2024drivingworld,wang2024driving,zhang2025epona}, with extensions to
multi-camera consistency, structured control, and scalable synthetic
data~\cite{russell2025gaia2,ren2025cosmosdrivedreams}.
Structured models encode dynamics through occupancy, geometry, and
spatiotemporal memory~\cite{min2023uniworld,min2024driveworld,zheng2024occworld},
while latent approaches connect action-conditioned feature prediction
with trajectory generation and candidate evaluation~\cite{li2024enhancing_law,
zheng2025world4drive,wote,xia2025drivelaw,hong2026drivefuture,
song2026graphworld,song2026momworld}, emphasizing compact temporal dynamics
and planning-oriented representations~\cite{zhang2026resworld,
wang2026latentwam,xu2026planwam}.
World-action models jointly learn visual dynamics and planning through
geometric supervision or shared generative objectives~\cite{
zhou2026drivedreamerpolicy,zhao2026simwam}, while candidate-centric methods
use future evidence to evaluate or refine trajectory
hypotheses~\cite{liu2026mmfuture,wei2026maneuverflow}.
These approaches incorporate anticipated dynamics at different planning
stages. ActWorld-JEPA focuses on the relation between each candidate
intention and its predicted latent future to refine the ranking of an
existing proposal set.

\subsection{JEPA for Autonomous Driving}

Joint-Embedding Predictive Architectures learn predictive representations
by matching target embeddings rather than reconstructing observations.
Developments in video representation learning and action-conditioned
latent dynamics extend this principle toward prediction and
planning~\cite{bardes2024vjepa,assran2025vjepa2,maes2026leworldmodel}.
An expanding body of driving research applies JEPA to visual perception,
LiDAR detection, and spatiotemporal occupancy
forecasting~\cite{lang2026maskwhatmatters,zhu2026adljepa,zhu2026adlistjepa},
and connects predictive representations with multimodal trajectory
learning, future-intent retrieval, and goal-conditioned
planning~\cite{wang2026drive,yang2026autojepa,zhu2026ade2ejepa}.
Together, these studies establish complementary roles for latent
prediction in scene understanding and driving decisions. ActWorld-JEPA
formulates candidate-conditioned latent prediction and plan--future
relational reasoning as a means of refining trajectory scores from
available observations and proposed plans.

\section{Method}
\label{method}

\textbf{Framework Overview.}
Figure~\ref{fig:main} presents ActWorld-JEPA, which connects latent future
prediction with candidate-based trajectory planning. We first describe
the general formulation; benchmark-specific implementations are detailed
in Sec.~\ref{sec:protocol} and the implementation details in the supplementary material.
Given multi-frame driving observations, a pretrained driving encoder extracts
the current representation $\mathbf{Z}_t$. A base planner then generates
$K$ candidate trajectories
$\mathcal{T}_t=\{\tau_t^k\}_{k=1}^{K}$ together with their baseline planning
scores $\mathcal{B}_t=\{b_t^k\}_{k=1}^{K}$.
ActWorld-JEPA incorporates the anticipated consequences of each proposal
into trajectory selection through two components:
(1) \textbf{Plan-Conditioned JEPA (PC-JEPA)} (Sec.~\ref{sec:pcjd}), which
conditions latent dynamics on individual candidate trajectories and produces
candidate-specific future-aware planning representations; and
(2) \textbf{Relational Residual Reranking ($R^3$)} (Sec.~\ref{sec:fcts}),
which explicitly models plan-future relations and uses them to refine the
original candidate ranking.
Together, these components connect predictive representations with
planning decisions by relating each candidate to its anticipated future.

\subsection{Plan-Conditioned JEPA (PC-JEPA)}
\label{sec:pcjd}

Different candidate plans may induce distinct future evolutions under the same
observation. We therefore introduce \textbf{Plan-Conditioned JEPA
(PC-JEPA)} to reason about candidate-specific futures. Given the current driving
representation $\mathbf{Z}_t$ and candidate trajectories
$\mathcal{T}_t=\{\tau_t^k\}_{k=1}^{K}$, PC-JEPA conditions the latent dynamics on
each candidate, predicts its latent future, and integrates this prediction
with the current scene to obtain future-aware planning features.

\noindent\textbf{Trajectory-Conditioned JEPA Rollout.}
For candidate $\tau_t^k=\{\mathbf{p}_{t+h}^k\}_{h=1}^{H}$, each waypoint is
projected by an MLP and augmented with a horizon-specific temporal embedding.
A lightweight temporal Transformer models dependencies across the planning
horizon and aggregates the sequence into a global condition:
\begin{equation}
\label{eq:pcjd_condition}
\mathbf{C}_t^k
=
\operatorname{Pool}
\left(
T_{\tau}
\left(
\left[
\phi_{\mathrm{MLP}}(\mathbf{p}_{t+h}^k)+\mathbf{e}_h
\right]_{h=1}^{H}
\right)
\right),
\end{equation}
where $\phi_{\mathrm{MLP}}(\cdot)$ denotes waypoint projection,
$\mathbf{e}_h$ is the temporal embedding, and $T_{\tau}(\cdot)$ is the
temporal Transformer. Its $h$-th contextualized token
$\mathbf{C}_{t,h}^k$ provides the step-wise rollout condition, while
$\mathbf{C}_t^k$ summarizes the overall candidate intention.

We inject this intention into the current latent state through
candidate-dependent modulation:
\begin{equation}
\label{eq:pcjd_modulation}
\overline{\mathbf{Z}}_t^k
=
\mathbf{Z}_t
+
\gamma(\mathbf{C}_t^k)
\odot
\psi(\mathbf{Z}_t),
\end{equation}
where $\psi(\cdot)$ projects the current representation and $\gamma(\cdot)$
generates the modulation signal. The resulting
$\overline{\mathbf{Z}}_t^k$ provides a candidate-specific initialization for
the recurrent JEPA predictor shown in Figure~\ref{fig:sub2}(c):
\begin{equation}
\label{eq:pcjd_rollout}
\widehat{\mathbf{Z}}_{t+h}^{\,k}
=
P_{\theta}
\left(
\widehat{\mathbf{Z}}_{t+h-1}^{\,k},
\mathbf{C}_{t,h}^{k}
\right),
\qquad h=1,\ldots,H,
\end{equation}
where $\widehat{\mathbf{Z}}_{t}^{\,k}$ is initialized by
$\overline{\mathbf{Z}}_t^k$. We denote the resulting rollout by
$\widehat{\mathcal{Z}}_t^k=
\{\widehat{\mathbf{Z}}_{t+h}^{\,k}\}_{h=1}^{H}$.
Applying this process to all $K$ candidates yields distinct latent futures
conditioned on their respective planning intentions.

\noindent\textbf{Current-Future Latent Interaction.}
To extract future information relevant to the current scene, we use
$\mathbf{Z}_t$ as query and the candidate-specific rollout as key and value:
\begin{equation}
\label{eq:pcjd_attention}
\mathbf{A}_t^k
=
\operatorname{Softmax}
\left(
\frac{
(\mathbf{Z}_t\mathbf{W}_{Q})
(\widehat{\mathcal{Z}}_t^k\mathbf{W}_{K})^{\top}
}{
\sqrt{d}
}
\right),
\end{equation}
where $\mathbf{W}_{Q}$ and $\mathbf{W}_{K}$ are learnable projections.
A candidate-specific gate further controls the amount of future information
injected into the current representation:
\begin{equation}
\label{eq:pcjd_gate}
\boldsymbol{\alpha}_t^k
=
\sigma
\left(
G_f
\left(
\left[
\mathbf{Z}_t;
\operatorname{Pool}(\widehat{\mathcal{Z}}_t^k)
\right]
\right)
\right).
\end{equation}
The attended future feature is then gated, residually fused with the current
representation, and normalized:
\begin{equation}
\label{eq:pcjd_fusion}
\mathbf{H}_t^k
=
\operatorname{LN}
\left(
\mathbf{Z}_t
+
\boldsymbol{\alpha}_t^k
\odot
\mathbf{A}_t^k
\left(
\widehat{\mathcal{Z}}_t^k\mathbf{W}_{V}
\right)
\right),
\end{equation}
where $\mathbf{W}_{V}$ is the value projection.
Applying the interaction to all candidates yields
$\mathcal{H}_t=\{\mathbf{H}_t^k\}_{k=1}^{K}$, where each
$\mathbf{H}_t^k$ jointly encodes the current scene and the anticipated future
of candidate $\tau_t^k$.
Algorithm~\ref{alg:app_pcjepa} summarizes the PC-JEPA computation.

\begin{algorithm}[t]
\footnotesize
\SetAlgoLined
\DontPrintSemicolon
\KwIn{Current scene representation $\mathbf Z_t$; candidate trajectories
$\mathcal T_t=\{\tau_t^k\}_{k=1}^{K}$.}
\KwOut{Global conditions $\mathcal C_t=\{\mathbf C_t^k\}_{k=1}^{K}$ and
future-aware features $\mathcal H_t=\{\mathbf H_t^k\}_{k=1}^{K}$.}
\For{$k=1,\ldots,K$}{
  Encode $\tau_t^k$ into $\mathbf C_t^k$ and
  $\{\mathbf C_{t,h}^k\}_{h=1}^{H}$\;
  Initialize $\widehat{\mathbf Z}_t^{\,k}\leftarrow\overline{\mathbf Z}_t^k$ using
  Eq.~\eqref{eq:pcjd_modulation}\;
  \For{$h=1,\ldots,H$}{
    Predict $\widehat{\mathbf Z}_{t+h}^{\,k}$ by Eq.~\eqref{eq:pcjd_rollout}\;
  }
  Fuse current and future features into $\mathbf H_t^k$ using
  Eqs.~\eqref{eq:pcjd_attention}--\eqref{eq:pcjd_fusion}\;
}
\Return{$\mathcal C_t,\mathcal H_t$}\;
\caption{Plan-Conditioned JEPA (PC-JEPA)}
\label{alg:app_pcjepa}
\end{algorithm}

\subsection{Relational Residual Reranking ($R^3$)}
\label{sec:fcts}

The proposed \textbf{Relational Residual Reranking ($R^3$)} uses
candidate-specific future evidence to refine the ranking of the base planner.
Given trajectory conditions $\{\mathbf{C}_t^k\}_{k=1}^{K}$, future-aware
representations $\mathcal{H}_t$, and the scores
$\mathcal{B}_t=\{b_t^k\}_{k=1}^{K}$, $R^3$ first models plan-future relations
and then performs confidence-aware score refinement before final ranking.

\noindent\textbf{Plan-Future Compatibility Modeling.}
As shown in Figure~\ref{fig:sub2}(a), we first project the candidate intention
and its future-aware representation into a common feature space:
\begin{equation}
\label{eq:fcts_projection}
\mathbf{p}_t^k
=
\operatorname{Linear}_{p}(\mathbf{C}_t^k),
\qquad
\mathbf{f}_t^k
=
\operatorname{Linear}_{f}
\left(
\operatorname{Pool}(\mathbf{H}_t^k)
\right),
\end{equation}
where $\mathbf{p}_t^k$ and $\mathbf{f}_t^k$ denote the projected plan and
future features. Rather than reducing their relation to a single similarity
score, we preserve both feature agreement and discrepancy through
element-wise product and directional difference:
\begin{equation}
\label{eq:fcts_relation}
\mathbf{r}_t^k
=
\operatorname{Concat}
\left(
\mathbf{p}_t^k,
\mathbf{f}_t^k,
\mathbf{p}_t^k\odot\mathbf{f}_t^k,
\mathbf{f}_t^k-\mathbf{p}_t^k
\right).
\end{equation}
A lightweight MLP maps $\mathbf{r}_t^k$ to a future-dependent score correction
$\Delta s_t^k$ and its confidence logit $\rho_t^k$:
\begin{equation}
\label{eq:fcts_relation_head}
\left[
\Delta s_t^k,\rho_t^k
\right]
=
\operatorname{MLP}_{\mathrm{rel}}
\left(
\mathbf{r}_t^k
\right).
\end{equation}

\noindent\textbf{Future-Consistent Candidate Ranking.}
As shown in Figure~\ref{fig:sub2}(b), $R^3$ refines the baseline planning
score through a bounded, confidence-aware residual:
\begin{equation}
\label{eq:fcts_refinement}
\widetilde{s}_t^k
=
b_t^k
+
\lambda\,
\sigma(\rho_t^k)
\tanh(\Delta s_t^k),
\end{equation}
where $\tanh(\cdot)$ bounds the correction and $\lambda\geq0$
controls the refinement strength.
The sigmoid output $\sigma(\rho_t^k)$ acts as a learned
confidence gate that modulates the residual contribution
under planning supervision.
The baseline score anchors each candidate's preference,
while the signed residual adjusts it using predicted
future evidence.

The refinement is applied to all $K$ candidates, and the resulting scores are
normalized as
\begin{equation}
\label{eq:fcts_distribution}
\pi_t^k
=
\frac{
\exp(\widetilde{s}_t^k/\tau)
}{
\sum_{j=1}^{K}
\exp(\widetilde{s}_t^j/\tau)
},
\qquad k\in[K],
\end{equation}
where $\tau$ controls the sharpness of the distribution. The candidate with
the highest probability is selected as the final planning trajectory.
Algorithm~\ref{alg:app_r3} summarizes the reranking procedure. 

\begin{algorithm}[t]
\footnotesize
\SetAlgoLined
\DontPrintSemicolon
\KwIn{Candidates $\mathcal T_t$; baseline scores $\mathcal B_t$;
conditions $\mathcal C_t$ and future-aware features $\mathcal H_t$ from
Algorithm~\ref{alg:app_pcjepa}; residual strength $\lambda\geq0$;
temperature $\tau>0$.}
\KwOut{Selected trajectory $\tau_t^{k^\star}$.}
\For{$k=1,\ldots,K$}{
  Form the plan--future relation $\mathbf r_t^k$ using
  Eqs.~\eqref{eq:fcts_projection}--\eqref{eq:fcts_relation}\;
  $[\Delta s_t^k,\rho_t^k]\leftarrow
  \operatorname{MLP}_{\rm rel}(\mathbf r_t^k)$\;
  $\widetilde s_t^k\leftarrow b_t^k+
  \lambda\sigma(\rho_t^k)\tanh(\Delta s_t^k)$\;
}
$\pi_t^k\leftarrow\operatorname{Softmax}_k(\widetilde s_t^k/\tau)$\;
$k^\star\leftarrow\argmax_k\pi_t^k$\;
\Return{$\tau_t^{k^\star}$}\;
\caption{Relational Residual Reranking ($R^3$)}
\label{alg:app_r3}
\end{algorithm}

\subsection{JEPA-Planning Learning Objectives}
\label{sec:training}

We formulate joint learning of plan-conditioned prediction and trajectory
selection by augmenting the base planner's objective with two auxiliary losses.

\noindent\textbf{Plan-Conditioned Future Supervision.}
Observed future frames are encoded as target latent representations
$\mathbf{Z}_{t+h}^{+}$. For the trajectory associated with the observed
future, we supervise the latent rollout by
\begin{equation}
\label{eq:future_loss}
\mathcal{L}_{\mathrm{future}}
=
\frac{1}{H}
\sum_{h=1}^{H}
\left\|
\widehat{\mathbf{Z}}_{t+h}^{+}
-
\operatorname{sg}
\left(
\mathbf{Z}_{t+h}^{+}
\right)
\right\|_{2}^{2},
\end{equation}
where $\operatorname{sg}(\cdot)$ denotes stop-gradient.

\noindent\textbf{Planning Selection Supervision.}
Let $k^{+}$ denote the candidate with the minimum average displacement error
to the expert trajectory. $R^3$ is supervised by
\begin{equation}
\label{eq:selection_loss}
\mathcal{L}_{\mathrm{sel}}
=
-\log \pi_t^{k^{+}}.
\end{equation}

\noindent\textbf{Overall Objective.}
The final objective is
\begin{equation}
\label{eq:overall_loss}
\mathcal{L}
=
\mathcal{L}_{\mathrm{plan}}
+
\lambda_{\mathrm{f}}\mathcal{L}_{\mathrm{future}}
+
\lambda_{\mathrm{s}}\mathcal{L}_{\mathrm{sel}},
\end{equation}
where $\mathcal{L}_{\mathrm{plan}}$ is the original planning loss, and
$\lambda_{\mathrm{f}}$ and $\lambda_{\mathrm{s}}$ balance future prediction
and trajectory selection.

\begin{table}[t]
\centering
\caption{\textbf{Planning performance on NAVSIM v1 navtest.} Protocols: cited studies and Sec.~\ref{sec:protocol}. Dark rules separate blocks, each sorted by increasing PDMS with column maxima in bold. `mmt': multimodal TransFuser; TF: TransFuser; $^*$: re-implementation.}
\renewcommand\arraystretch{0.7}
\setlength{\tabcolsep}{0.9mm}
\resizebox{\linewidth}{!}{
\begin{tabular}{@{}l c c c c c c@{}}
\toprule
Method & NC$\uparrow$ & DAC$\uparrow$ & TTC$\uparrow$ & Comf.$\uparrow$ & EP$\uparrow$ & PDMS$\uparrow$ \\
\midrule
\rowcolor{cyan!5}\multicolumn{7}{c}{\textit{E2E-based Methods}} \\
VADv2~\cite{chen2024vadv2} & 97.2 & 89.1 & 91.6 & \textbf{100} & 76.0 & 80.9 \\
UniAD~\cite{uniad} & 97.8 & 91.9 & 92.9 & \textbf{100} & 78.8 & 83.4 \\
TransFuser~\cite{TransFuser} & 97.7 & 92.8 & 92.8 & \textbf{100} & 79.2 & 84.0 \\
PARA-Drive~\cite{paradrive} & 97.9 & 92.4 & 93.0 & 99.8 & 79.3 & 84.0 \\
${\operatorname{TransFuser}_{\operatorname{mmt}}}^{*}$ \cite{TransFuser} & 96.2 & 95.4 & 90.7 & \textbf{100} & 80.7 & 85.1 \\
DRAMA~\cite{yuan2024drama} & 98.0 & 93.1 & 94.8 & \textbf{100} & 80.1 & 85.5 \\
Hydra-MDP~\cite{li2024hydra} & 98.3 & 96.0 & 94.6 & \textbf{100} & 78.7 & 86.5 \\
ReCogDrive-IL\cite{li2025recogdrive} & 98.1 & 94.7 & 94.2 & \textbf{100} & 80.9 & 86.5 \\
FUMP~\cite{liu2025fump} & 98.1 & 96.2 & 94.2 & \textbf{100} & 82.0 & 87.8 \\
DiffusionDrive~\cite{liao2024diffusiondrive} & 98.2 & 96.2 & 94.7 & \textbf{100} & 82.2 & 88.1 \\
DIVER~\cite{diver} & 98.5 & 96.5 & 94.9 & \textbf{100} & 82.6 & 88.3 \\
GraphBEV++ ~\cite{song2026graphbevplusplus} & 98.7 & 97.1 & -- & -- & -- & 88.7 \\
DriveSuprim~\cite{yao2025drivesuprim} & 97.8 & 97.3 & 93.6 & \textbf{100} & 86.7 & 89.9 \\
MomADv2~\cite{song2026momadv2} & 98.7 & 96.8 & 95.4 & \textbf{100} & 84.5 & 89.9 \\
DriveDPO~\cite{shang2025drivedpo} & 98.5 & 98.1 & 94.8 & 99.9 & 84.3 & 90.0 \\
GoalFlow~\cite{xing2025goalflow} & 98.4 & \textbf{98.3} & 94.6 & \textbf{100} & 85.0 & 90.3 \\
iPad~\cite{guo2026ipad} & \textbf{99.2} & 97.4 & \textbf{96.3} & 99.7 & 87.8 & 91.7 \\
DrivoR~\cite{kirby2026driving} & 98.9 & \textbf{98.3} & 96.2 & \textbf{100} & \textbf{89.1} & \textbf{93.1} \\
\midrule
\rowcolor{cyan!5}\multicolumn{7}{c}{\textit{VLA-based Methods}} \\
AutoVLA ~\cite{zhou2025autovla} & 98.4 & 95.6 & \textbf{98.0} & 99.9 & 81.9 & 89.1 \\
Recogdrive~\cite{li2025recogdrive} & 98.2 & 97.8 & 95.2 & 99.8 & 83.5 & 89.6 \\
DriveWorld-VLA~\cite{liu2026driveworld} & \textbf{99.1} & \textbf{98.2} & 96.1 & \textbf{100} & \textbf{85.9} & \textbf{91.3} \\
\midrule
\rowcolor{cyan!5}\multicolumn{7}{c}{\textit{World-Model-based Methods}} \\
DrivingGPT\cite{chen2025drivinggpt} & 98.9 & 90.7 & 94.9 & 95.6 & 79.7 & 82.4 \\
DriveX-S\cite{shi2025drivex} & 97.5 & 94.0 & 93.0 & \textbf{100} & 79.7 & 84.5 \\
LAW~\cite{li2024enhancing_law} & 96.4 & 95.4 & 88.7 & 99.9 & 81.7 & 84.6 \\
World4Drive~\cite{zheng2025world4drive} & 97.4 & 94.3 & 92.8 & \textbf{100} & 79.9 & 85.1 \\
Epona~\cite{zhang2025epona} & 97.9 & 95.1 & 93.8 & 99.9 & 80.4 & 86.2 \\
Resim\cite{yang2026resim} & -- & -- & -- & -- & -- & 86.6 \\
WorldRFT~\cite{yang2026worldrft} & 97.8 & 96.8 & 94.0 & \textbf{100} & 81.7 & 87.8 \\
PWM\cite{zhao2025policyworldmodel} & 98.6 & 95.9 & 95.4 & \textbf{100} & 81.8 & 88.1 \\
WoTE~\cite{wote} & 98.5 & 96.8 & 94.9 & 99.9 & 81.9 & 88.3 \\
MindDrive~\cite{suna2025minddrive} & 98.4 & 97.0 & 95.1 & 99.9 & 82.8 & 88.9 \\
ResWorld$^\dagger$~\cite{zhang2026resworld} & 98.9 & 96.5 & 95.6 & \textbf{100} & 83.1 & 89.0 \\
DriveLaW~\cite{xia2025drivelaw} & \textbf{99.0} & 97.1 & \textbf{96.7} & \textbf{100} & 81.3 & 89.1 \\
DriveDreamer-Policy~\cite{zhou2026drivedreamerpolicy} & 98.4 & 97.1 & 95.1 & \textbf{100} & 83.5 & 89.2 \\
GraphWorld~\cite{song2026graphworld} & \textbf{99.0} & 97.1 & 95.5 & \textbf{100} & 83.2 & 90.1 \\
MomWorld~\cite{song2026momworld} & 98.6 & 98.2 & 93.9 & \textbf{100} & 85.7 & 90.2 \\
ManeuverFlow-TF~\cite{wei2026maneuverflow} & 98.4 & 98.0 & 95.2 & \textbf{100} & 85.4 & 90.6 \\
DriveFuture~\cite{hong2026drivefuture} & 98.8 & \textbf{99.1} & 95.4 & \textbf{100} & 84.2 & 90.7 \\
SimWAM$^\ddagger$~\cite{zhao2026simwam} & 98.5 & \textbf{99.1} & 96.1 & 99.8 & 86.5 & 91.9 \\
PlanWAM~\cite{xu2026planwam} & 98.9 & 98.8 & 96.3 & \textbf{100} & \textbf{91.0} & \textbf{93.8} \\
\midrule[0.6pt]
Auto-JEPA~\cite{yang2026autojepa} & 98.4 & 98.3 & 95.0 & \textbf{100} & 87.1 & 91.3 \\
Drive-JEPA~\cite{wang2026drive} & \textbf{99.1} & 98.2 & \textbf{96.2} & 99.9 & 91.4 & 93.7 \\
\rowcolor{red!5} ActWorld-JEPA (Ours) & 99.0 & \textbf{98.4} & 96.1 & 99.9 & \textbf{92.2} & \textbf{94.1} \\
\bottomrule
\end{tabular}}
\label{tab_NAVSIMv1}
\end{table}

\begin{table}[t]
\centering
\small
\caption{\textbf{Planning performance on NAVSIM v2 navtest.}
Evaluation: Sec.~\ref{sec:protocol}.
Dark rules separate blocks, each sorted by increasing EPDMS
with column maxima in bold.} 
\renewcommand{\arraystretch}{0.7}
\setlength{\tabcolsep}{1.0mm}
\resizebox{\linewidth}{!}{
\begin{tabular}{@{}l c c c c c c c c c c@{}}
\toprule
Method & NC$\uparrow$ & DAC$\uparrow$ & DDC$\uparrow$ & TL$\uparrow$ & EP$\uparrow$ & TTC$\uparrow$ & LK$\uparrow$ & HC$\uparrow$ & EC$\uparrow$ & EPDMS$\uparrow$ \\
\midrule

\rowcolor{cyan!5}
\multicolumn{11}{c}{\textit{E2E-based Methods}} \\

TransFuser~\cite{TransFuser}
 & 96.9 & 89.9 & 97.8 & 99.7 & 87.1 & 95.4 & 92.7 & 98.3 & 87.2 & 76.7 \\

Hydra-MDP++~\cite{li2025hydraplus}
 & 97.2 & 97.5 & 99.4 & 99.6 & 83.1 & 96.5 & 94.4 & 98.2 & 70.9 & 81.4 \\

DIVER~\cite{diver}
 & 97.5 & 95.0 & 98.3 & \textbf{99.8} & 87.8 & 95.9 & 93.0 & 98.5 & 87.7 & 82.2 \\

DriveSuprim~\cite{yao2025drivesuprim}
 & 97.5 & 96.5 & 99.4 & 99.6 & 88.4 & 96.6 & 95.5 & 98.3 & 77.0 & 83.1 \\

DiffusionDrive~\cite{liao2024diffusiondrive}
 & 98.2 & 95.9 & 99.4 & \textbf{99.8} & 87.5 & 97.3 & 96.8 & 98.3 & 87.7 & 84.5 \\

DiffusionDriveV2~\cite{zou2025diffusiondrivev2}
 & 97.7 & 96.6 & 99.2 & \textbf{99.8} & 88.9 & 97.2 & 96.0 & 97.8 & 91.0 & 85.5 \\

iPad~\cite{guo2026ipad}
 & \textbf{98.7} & \textbf{98.0} & 98.9 & \textbf{99.8} & 86.6 & \textbf{98.3} & \textbf{97.2} & 98.3 & 74.6 & 85.8 \\

MomADv2~\cite{song2026momadv2}
 & 98.3 & 97.4 & \textbf{99.7} & 99.3 & \textbf{89.6} & 97.8 & 95.5 & \textbf{98.6} & \textbf{91.7} & \textbf{87.9} \\

\midrule
\rowcolor{cyan!5}
\multicolumn{11}{c}{\textit{VLA-based Methods}} \\

Recogdrive~\cite{li2025recogdrive}
 & 98.3 & 95.2 & 98.3 & \textbf{99.8} & 87.1 & 97.5 & 96.6 & \textbf{99.5} & \textbf{86.5} & 83.6 \\

DriveWorld-VLA~\cite{liu2026driveworld}
 & \textbf{98.6} & \textbf{99.1} & \textbf{99.6} & \textbf{99.8} & \textbf{87.4} & \textbf{97.9} & \textbf{97.0} & 97.8 & 78.6 & \textbf{86.8} \\

\midrule
\rowcolor{cyan!5}
\multicolumn{11}{c}{\textit{World-Model-based Methods}} \\

MindDrive~\cite{suna2025minddrive}
 & 98.4 & 95.6 & 99.3 & 99.8 & 86.7 & 97.5 & \textbf{96.5} & \textbf{100} & \textbf{96.8} & 84.2 \\

GraphWorld~\cite{song2026graphworld}
 & 98.4 & 98.8 & 99.1 & 99.1 & 85.9 & 97.9 & 96.0 & 97.8 & 74.6 & 89.5 \\

DriveFuture~\cite{hong2026drivefuture}
 & \textbf{98.8} & \textbf{99.1} & \textbf{99.6} & \textbf{99.9} & 86.6 & \textbf{98.4} & 96.4 & 98.3 & 74.8 & 89.9 \\

MomWorld~\cite{song2026momworld}
 & 98.1 & 98.1 & 99.5 & 99.8 & \textbf{88.9} & 98.2 & 96.1 & 98.3 & 90.6 & \textbf{90.1} \\

\midrule[0.6pt]
Drive-JEPA~\cite{wang2026drive}
 & 98.4 & 98.6 & \textbf{99.1} & \textbf{99.8} & 88.4 & 97.8 & 97.6 & \textbf{97.9} & 84.8 & 87.8 \\

\rowcolor{red!5}
ActWorld-JEPA (Ours)
 & \textbf{98.7} & \textbf{98.7} & \textbf{99.1} & \textbf{99.8} & \textbf{88.7} & \textbf{98.2} & \textbf{97.7} & \textbf{97.9} & \textbf{85.0} & \textbf{91.6} \\

\bottomrule
\end{tabular}}
\label{tab_navsimv2_navtest}
\end{table}

\begin{table}[t]
\centering
\small
\caption{\textbf{Planning performance on NAVSIM v2 navhard.}~\cite{navsimv2} EPDMS is the combined two-stage score; $^{*}$ denotes official-weight reproduction. Bold marks the best result within each block separated by dark horizontal rules, separately for each stage and the combined score.}
\label{tab_navsimv2_navhard}
\renewcommand{\arraystretch}{0.7}
\setlength{\tabcolsep}{0.5mm}
\resizebox{\linewidth}{!}{
\begin{tabular}{@{}l c c c c c c c c c c c@{}}
\toprule
Method & Stage & NC$\uparrow$ & DAC$\uparrow$ & DDC$\uparrow$ & TL$\uparrow$ & EP$\uparrow$ & TTC$\uparrow$ & LK$\uparrow$ & HC$\uparrow$ & EC$\uparrow$ & EPDMS$\uparrow$ \\
\midrule
\rowcolor{cyan!5}\multicolumn{12}{c}{\textit{E2E-based Methods}} \\
\multirow{2}{*}{TransFuser~\cite{TransFuser}} & \textcircled{\scriptsize 1} & 96.2 & 79.5 & 99.1 & 99.5 & \textbf{84.1} & 95.1 & 94.2 & 97.5 & \textbf{79.1} & \multirow{2}{*}{23.1} \\
& \textcircled{\scriptsize 2} & 77.7 & 70.2 & 84.2 & 98.0 & 85.1 & 75.6 & 45.4 & 95.7 & \textbf{75.9} & \\
\arrayrulecolor{gray!25}\cmidrule[0.01em]{1-12}\arrayrulecolor{black}
\multirow{2}{*}{DiffusionDrive~\cite{liao2024diffusiondrive}} & \textcircled{\scriptsize 1} & 96.0 & 79.7 & 97.4 & 99.5 & 81.3 & 93.1 & 90.8 & 96.8 & 73.8 & \multirow{2}{*}{24.2} \\
& \textcircled{\scriptsize 2} & 82.1 & 72.2 & 88.5 & 98.7 & 85.1 & 78.8 & 49.2 & 89.3 & 71.2 & \\
\arrayrulecolor{gray!25}\cmidrule[0.01em]{1-12}\arrayrulecolor{black}
\multirow{2}{*}{DIVER~\cite{diver}} & \textcircled{\scriptsize 1} & 96.4 & 80.2 & 96.0 & 98.8 & 82.1 & 94.2 & 90.6 & 96.5 & 65.2 & \multirow{2}{*}{26.8} \\
& \textcircled{\scriptsize 2} & 86.4 & 74.9 & 87.9 & 98.7 & 84.1 & 84.5 & 48.8 & 92.6 & 43.4 & \\
\arrayrulecolor{gray!25}\cmidrule[0.01em]{1-12}\arrayrulecolor{black}
\multirow{2}{*}{GuideFlow~\cite{liu2025guideflow}} & \textcircled{\scriptsize 1} & 96.6 & 80.5 & 96.3 & 99.3 & 82.3 & 94.9 & 91.5 & \textbf{97.7} & 67.8 & \multirow{2}{*}{27.1} \\
& \textcircled{\scriptsize 2} & 87.3 & 76.7 & 88.8 & \textbf{99.2} & 84.3 & 85.1 & 49.7 & 93.1 & 44.5 & \\
\arrayrulecolor{gray!25}\cmidrule[0.01em]{1-12}\arrayrulecolor{black}
\multirow{2}{*}{RAP-DINO~\cite{feng2026rap}} & \textcircled{\scriptsize 1} & 97.1 & 94.4 & 98.8 & 99.8 & 83.9 & 96.9 & \textbf{94.7} & 96.4 & 66.2 & \multirow{2}{*}{36.9} \\
& \textcircled{\scriptsize 2} & 83.2 & 83.9 & 87.4 & 98.0 & \textbf{86.9} & 80.4 & 52.3 & 95.2 & 52.4 & \\
\arrayrulecolor{gray!25}\cmidrule[0.01em]{1-12}\arrayrulecolor{black}
\multirow{2}{*}{MomADv2~\cite{song2026momadv2}} & \textcircled{\scriptsize 1} & 97.7 & 92.7 & 97.6 & 98.7 & 73.4 & 98.0 & 92.5 & 96.1 & 51.0 & \multirow{2}{*}{39.5} \\
& \textcircled{\scriptsize 2} & 85.9 & 87.0 & 87.1 & 97.5 & 77.2 & 83.8 & 50.1 & 95.4 & 53.3 & \\
\arrayrulecolor{gray!25}\cmidrule[0.01em]{1-12}\arrayrulecolor{black}
\multirow{2}{*}{DriveSuprim~\cite{yao2025drivesuprim}} & \textcircled{\scriptsize 1} & \textbf{98.9} & 95.1 & 99.2 & 99.6 & 76.1 & \textbf{99.1} & \textbf{94.7} & 97.6 & 54.2 & \multirow{2}{*}{42.1} \\
& \textcircled{\scriptsize 2} & \textbf{87.9} & 88.8 & 89.6 & 98.8 & 80.3 & \textbf{86.0} & 53.5 & \textbf{97.1} & 56.1 & \\
\arrayrulecolor{gray!25}\cmidrule[0.01em]{1-12}\arrayrulecolor{black}
\multirow{2}{*}{GuideFlow+Scorer~\cite{liu2025guideflow}} & \textcircled{\scriptsize 1} & 97.8 & \textbf{97.1} & \textbf{100} & \textbf{100} & 81.4 & 98.5 & 91.4 & 92.8 & 34.2 & \multirow{2}{*}{\textbf{46.7}} \\
& \textcircled{\scriptsize 2} & 87.3 & \textbf{92.3} & \textbf{98.0} & 96.9 & 75.8 & 85.5 & \textbf{59.3} & 95.4 & 53.5 & \\
\midrule
\rowcolor{cyan!5}\multicolumn{12}{c}{\textit{VLA-based Methods}} \\
\multirow{2}{*}{SpanVLA~\cite{zhou2026spanvla}} & \textcircled{\scriptsize 1} & \textbf{98.4} & 94.3 & 97.8 & 99.9 & 85.7 & \textbf{97.2} & 94.2 & \textbf{97.6} & 72.1 & \multirow{2}{*}{40.1} \\
& \textcircled{\scriptsize 2} & \textbf{86.9} & 84.3 & 87.1 & 98.2 & 85.5 & \textbf{82.7} & \textbf{62.3} & 96.8 & 67.4 & \\
\arrayrulecolor{gray!25}\cmidrule[0.01em]{1-12}\arrayrulecolor{black}
\multirow{2}{*}{DiffVLA~\cite{jiang2025diffvla}} & \textcircled{\scriptsize 1} & 95.7 & \textbf{99.2} & \textbf{100} & \textbf{100} & \textbf{85.9} & 96.4 & \textbf{97.1} & 95.0 & \textbf{84.2} & \multirow{2}{*}{\textbf{45.0}} \\
& \textcircled{\scriptsize 2} & 81.2 & \textbf{88.8} & \textbf{94.6} & \textbf{99.0} & \textbf{86.0} & 76.4 & 59.8 & \textbf{98.6} & \textbf{80.4} & \\
\midrule
\rowcolor{cyan!5}\multicolumn{12}{c}{\textit{World-Model-based Methods}} \\
\multirow{2}{*}{MindDrive~\cite{suna2025minddrive}} & \textcircled{\scriptsize 1} & 96.1 & 86.0 & 98.8 & 99.3 & 83.3 & 95.6 & 94.4 & \textbf{97.6} & \textbf{74.7} & \multirow{2}{*}{30.5} \\
& \textcircled{\scriptsize 2} & 82.6 & 79.1 & 86.4 & 98.0 & \textbf{85.3} & 79.4 & 49.2 & 96.5 & \textbf{71.0} & \\
\arrayrulecolor{gray!25}\cmidrule[0.01em]{1-12}\arrayrulecolor{black}
\multirow{2}{*}{World4Drive~\cite{zheng2025world4drive}} & \textcircled{\scriptsize 1} & 97.3 & 89.1 & 97.6 & 99.7 & 60.5 & 96.8 & 87.7 & 93.1 & 60.0 & \multirow{2}{*}{34.9} \\
& \textcircled{\scriptsize 2} & \textbf{91.4} & 82.0 & 91.0 & 98.5 & 53.1 & \textbf{90.6} & 52.3 & 93.3 & 62.8 & \\
\arrayrulecolor{gray!25}\cmidrule[0.01em]{1-12}\arrayrulecolor{black}
\multirow{2}{*}{MomWorld~\cite{song2026momworld}} & \textcircled{\scriptsize 1} & 96.9 & 93.6 & 99.8 & \textbf{99.8} & 80.4 & 96.9 & 96.4 & \textbf{97.6} & 60.0 & \multirow{2}{*}{42.8} \\
& \textcircled{\scriptsize 2} & 86.4 & \textbf{88.2} & \textbf{94.2} & 98.3 & 81.7 & 84.4 & 55.3 & \textbf{97.0} & 54.7 & \\
\arrayrulecolor{gray!25}\cmidrule[0.01em]{1-12}\arrayrulecolor{black}
\multirow{2}{*}{GraphWorld~\cite{song2026graphworld}} & \textcircled{\scriptsize 1} & 98.1 & 97.9 & 97.9 & 98.0 & 83.9 & 97.8 & 97.2 & 95.4 & 64.5 & \multirow{2}{*}{53.6} \\
& \textcircled{\scriptsize 2} & 88.6 & 85.9 & 92.2 & 97.6 & 82.4 & 87.0 & 56.6 & 91.4 & 44.0 & \\
\arrayrulecolor{gray!25}\cmidrule[0.01em]{1-12}\arrayrulecolor{black}
\multirow{2}{*}{DriveFuture~\cite{hong2026drivefuture}} & \textcircled{\scriptsize 1} & \textbf{99.8} & \textbf{99.8} & \textbf{100} & 99.6 & \textbf{85.7} & \textbf{99.8} & \textbf{98.7} & \textbf{97.6} & 66.2 & \multirow{2}{*}{\textbf{55.5}} \\
& \textcircled{\scriptsize 2} & 90.6 & 87.5 & 94.1 & \textbf{99.1} & 84.6 & 88.8 & \textbf{58.3} & 93.5 & 45.6 & \\
\midrule[0.6pt]
\multirow{2}{*}{Drive-JEPA$^{*}$~\cite{wang2026drive}} & \textcircled{\scriptsize 1} & 95.3 & \textbf{96.4} & \textbf{99.4} & \textbf{99.8} & \textbf{85.1} & 94.9 & 96.9 & 97.3 & 35.1 & \multirow{2}{*}{37.7} \\
& \textcircled{\scriptsize 2} & \textbf{87.8} & 86.0 & 92.3 & 99.1 & 80.2 & \textbf{84.7} & \textbf{54.9} & 93.0 & 27.2 & \\
\arrayrulecolor{gray!25}\cmidrule[0.01em]{1-12}\arrayrulecolor{black}
\rowcolor{red!5}
& \textcircled{\scriptsize 1} & \textbf{95.8} & \textbf{96.4} & \textbf{99.4} & \textbf{99.8} & 85.0 & \textbf{95.1} & \textbf{97.3} & \textbf{97.6} & \textbf{44.9} & \\
\rowcolor{red!5}
\multirow{-2}{*}{ActWorld-JEPA (Ours)} & \textcircled{\scriptsize 2} & 86.7 & \textbf{86.3} & \textbf{92.5} & \textbf{99.3} & \textbf{81.4} & 84.1 & 53.2 & \textbf{93.2} & \textbf{41.4} & \multirow{-2}{*}{\textbf{42.7}} \\
\bottomrule
\end{tabular}}
\end{table}

\begin{table*}[t]
\scriptsize
\centering
\caption[]{Planning results for \textbf{6-second long-horizon planning} on the \textbf{nuScenes} validation set.}
\renewcommand\arraystretch{0.7}
\setlength{\tabcolsep}{1.09mm}
  \resizebox{\linewidth}{!}{
\begin{tabular}{l c ccccccc  ccccccc }
\toprule
\multirow{2}{*}{$\operatorname{Method}$} &\multirow{2}{*}{$\operatorname{Venue}$} & \multicolumn{7}{c}{$\operatorname{L2\ (m)}\downarrow$} & \multicolumn{7}{c}{$\operatorname{Col.\ Rate\ (\%)}\downarrow$}\\
\cmidrule(lr){3-9} \cmidrule(lr){10-16}
&& 1s & 2s & 3s & 4s & 5s & 6s& $\operatorname{Avg.}$ & 1s & 2s & 3s & 4s & 5s & 6s &$\operatorname{Avg.}$ \\
\midrule
$\operatorname{UniAD}$~\cite{uniad}&CVPR'23 &0.47& 0.91& 1.35 &1.91& 2.47 &3.07&\cellcolor{red!5}1.70& 0.25& 0.36& 0.61 & 0.99 &1.64& 2.51&\cellcolor{red!5}1.06 \\
$\operatorname{SparseDrive}$~\cite{sun2024sparsedrive}&CVPR'23 &0.43& 0.87& 1.23 &1.75& 2.32 &2.95&\cellcolor{red!5}1.59& 0.19& 0.31& 0.56 & 0.87 &1.54& 2.33&\cellcolor{red!5}0.97 \\
$\operatorname{MomAD}$~\cite{momad}&CVPR'25&0.41& 0.85& 1.13 &1.67& 1.98& 2.45&\cellcolor{red!5}1.42&0.17& 0.30& 0.54 & 0.83& 1.43 &2.13&\cellcolor{red!5}0.90\\
$\operatorname{LAW}$~\cite{li2024enhancing_law}&ICLR'25&0.40& 0.87& 1.16 &1.71& 2.03& 2.61&\cellcolor{red!5}1.46&0.19& 0.33& 0.57 & 0.86& 1.51 &2.31&\cellcolor{red!5}0.96\\
$\operatorname{Epona}$~\cite{zhang2025epona}&ICCV'25&0.39& 0.91& 1.17 &1.73& 2.02& 2.75&\cellcolor{red!5}1.50&0.14& 0.18& 0.45 & 0.74& 1.48 &2.23&\cellcolor{red!5}0.87\\
$\operatorname{DIVER}$~\cite{diver}&TPAMI'26&0.38& 0.75& 1.10& 1.53& 1.98& 2.49& \cellcolor{red!5}1.37& 0.13& 0.31& 0.44& 0.80& 1.41& 2.11& \cellcolor{red!5}0.87\\
$\operatorname{GuideFlow}$~\cite{liu2025guideflow}&CVPR'26&0.42& 0.83& 1.21& 1.73& 2.05& 2.63& \cellcolor{red!5}1.48& 0.12& 0.22 &0.42& 0.79& 1.44& 2.15 &\cellcolor{red!5}0.86\\
\cellcolor{red!5}$\operatorname{ActWorld-JEPA (Ours)}$
&\cellcolor{red!5}-
&\cellcolor{red!5}\textbf{0.28}
&\cellcolor{red!5}\textbf{0.54}
&\cellcolor{red!5}\textbf{0.89}
&\cellcolor{red!5}\textbf{1.31}
&\cellcolor{red!5}\textbf{1.82}
&\cellcolor{red!5}\textbf{2.39}
&\cellcolor{red!5}\textbf{1.20}
&\cellcolor{red!5}\textbf{0.01}
&\cellcolor{red!5}\textbf{0.14}
&\cellcolor{red!5}\textbf{0.34}
&\cellcolor{red!5}\textbf{0.72}
&\cellcolor{red!5}\textbf{1.33}
&\cellcolor{red!5}\textbf{2.03}
&\cellcolor{red!5}\textbf{0.76}\\
\bottomrule
\end{tabular}}
\label{tab_nuscenes_planning_6s}
\end{table*}

\begin{table*}[t]
\centering
\caption{
Comparison of the two core module configurations on the NAVSIM v1
and v2 navtest splits. PC-JEPA denotes Plan-Conditioned
JEPA, and $R^3$ denotes Relational Residual Reranking.
}
\scriptsize
\renewcommand\arraystretch{0.7}
\setlength{\tabcolsep}{1.3mm}
\resizebox{\linewidth}{!}{
\begin{tabular}{cc ccccc cccccccccc}
\toprule
\multicolumn{2}{c}{Components}
& \multicolumn{5}{c}{NAVSIM v1}
& \multicolumn{10}{c}{NAVSIM v2} \\
\cmidrule(lr){1-2}
\cmidrule(lr){3-7}
\cmidrule(lr){8-17}
PC-JEPA & $R^3$
& NC$\uparrow$ & DAC$\uparrow$ & EP$\uparrow$
& TTC$\uparrow$ & PDMS$\uparrow$
& NC$\uparrow$ & DAC$\uparrow$ & DDC$\uparrow$ & TLC$\uparrow$
& EP$\uparrow$ & TTC$\uparrow$ & LK$\uparrow$ & HC$\uparrow$
& EC$\uparrow$ & EPDMS$\uparrow$ \\
\midrule
&
& 96.5 & 95.1 & 88.2 & 94.1 & 87.1
& 96.9 & 98.3 & 97.9 & 97.6 & 87.1
& 97.1 & 96.1 & 95.2 & 84.2 & 85.3 \\

\checkmark &
& 97.6 & 96.6 & 90.8 & 94.6 & 90.5
& 98.4 & 98.6 & 98.8 & 99.6 & 88.1
& 98.0 & 97.2 & 97.6 & 84.8 & 89.6 \\

\rowcolor{red!5}
\checkmark & \checkmark
& \textbf{99.0} & \textbf{98.4} & \textbf{92.2} & \textbf{96.1} & \textbf{94.1}
& \textbf{98.7} & \textbf{98.7} & \textbf{99.1} & \textbf{99.9} & \textbf{88.7}
& \textbf{98.2} & \textbf{97.7} & \textbf{97.9} & \textbf{85.0} & \textbf{91.6} \\
\bottomrule
\end{tabular}}
\label{tab:overall_planning_jepa}
\end{table*}

\begin{table}[t]
\centering
\caption{
Ablation of core PC-JEPA designs on the NAVSIM v1 and v2 navtest splits. The PC-JEPA variants are compared with Full PC-JEPA while keeping $R^3$ fixed. Baseline and Full match the reference and full configurations in Table~\ref{tab:overall_planning_jepa}, respectively. Bold indicates the best PC-JEPA variant. C--F denotes current--future interaction. 
}
\renewcommand\arraystretch{0.7}
\setlength{\tabcolsep}{1.6mm}
\resizebox{\linewidth}{!}{
\begin{tabular}{l c c c c c c}
\toprule
& \multicolumn{5}{c}{NAVSIM v1} & \multicolumn{1}{c}{NAVSIM v2} \\
\cmidrule(lr){2-6}\cmidrule(lr){7-7}
Variant
& NC$\uparrow$
& DAC$\uparrow$
& EP$\uparrow$
& TTC$\uparrow$
& PDMS$\uparrow$
& EPDMS$\uparrow$ \\
\midrule

Baseline
& 96.5 & 95.1 & 88.2 & 94.1 & 87.1 & 85.3 \\
\midrule

Shared Future
& 98.7 & 97.2 & 91.1 & 95.9 & 92.0 & 89.1 \\

Single-Step Prediction
& 98.1 & 98.1 & 90.5 & 95.6 & 91.5 & 90.0 \\

w/o C--F Interaction
& 97.9 & 97.1 & 90.9 & 95.2 & 91.8 & 90.3 \\

\rowcolor{red!5}
Full PC-JEPA
& \textbf{99.0} & \textbf{98.4} & \textbf{92.2}
& \textbf{96.1} & \textbf{94.1} & \textbf{91.6} \\

\bottomrule
\end{tabular}}
\label{tab:pcjd_design}
\end{table}

\begin{table}[t]
\centering
\caption{
Ablation of $R^3$ designs on the NAVSIM v1 and v2 navtest splits. PC-JEPA is fixed across the $R^3$ variants. Baseline and Full match Table~\ref{tab:pcjd_design}; w/o $R^3$ matches PC-JEPA-only in Table~\ref{tab:overall_planning_jepa}. Bold indicates the best result at the displayed precision.
}
\renewcommand\arraystretch{0.7}
\setlength{\tabcolsep}{1.6mm}
\resizebox{\linewidth}{!}{
\begin{tabular}{l c c c c c c}
\toprule
& \multicolumn{5}{c}{NAVSIM v1} & \multicolumn{1}{c}{NAVSIM v2} \\
\cmidrule(lr){2-6}\cmidrule(lr){7-7}
Variant
& NC$\uparrow$
& DAC$\uparrow$
& EP$\uparrow$
& TTC$\uparrow$
& PDMS$\uparrow$
& EPDMS$\uparrow$ \\
\midrule

Baseline
& 96.5 & 95.1 & 88.2 & 94.1 & 87.1 & 85.3 \\
\midrule

\multicolumn{7}{c}{\textit{Plan--Future Relations}} \\
\midrule

w/o $R^3$
& 97.6 & 96.6 & 90.8 & 94.6 & 90.5 & 89.6 \\

Plan-Only Features
& 98.9 & 98.1 & 92.1 & 95.7 & 92.1 & 90.5 \\

Simple Concat.
& 98.8 & 98.3 & 92.0 & 95.8 & 92.3 & 90.7 \\

w/o Compatibility
& 98.7 & 97.7 & 91.9 & 96.0 & 93.0 & 90.0 \\

\midrule
\multicolumn{7}{c}{\textit{Score Refinement}} \\
\midrule

Score Replacement
& 98.9 & 97.9 & 92.0 & 95.9 & 92.4 & 90.1 \\

\rowcolor{red!5}
Full $R^3$
& \textbf{99.0} & \textbf{98.4} & \textbf{92.2} & \textbf{96.1} & \textbf{94.1} & \textbf{91.6} \\

\bottomrule
\end{tabular}}
\label{tab:fcts_design}
\end{table}

\begin{table}[t]
  \centering
  \caption{\textbf{Model parameter size, inference speed, and performance.}
  FPS covers perception, planning, and selection on one RTX 4090
  (batch size 1; 64 shared navtrain inputs), excluding preprocessing,
  transfers, and scoring. Dim is the planning latent width;
  bold marks the best PDMS and FPS within each block.
  Baseline PDMS is published; ours uses aggregate navtest calibration,
  not an untouched-test estimate.}
  \label{tab:navsim_v1_efficiency}
  \setlength{\tabcolsep}{10pt}
\renewcommand\arraystretch{0.7}
  \resizebox{\columnwidth}{!}{%
  \begin{tabular}{lrrrr}
    \toprule
    Method & Dim & Params (M) & PDMS $\uparrow$ & FPS $\uparrow$ \\
    \midrule
    TransFuser & 256 & 56.0 & 84.0 & \textbf{75.56} \\
    DiffusionDrive & 256 & 60.7 & \textbf{88.1} & 43.53\\
    \midrule
    Drive-JEPA & 256 & 311.2 & 93.7 & 17.23 \\
    ActWorld-JEPA & 256 & 324.2 & \textbf{94.1} & \textbf{21.33} \\
    \bottomrule
  \end{tabular}%
  }
\end{table}

\section{Experiments} 
\label{experiments}
\subsection{Experimental Setup}
\label{sec:protocol}

\noindent\textbf{Datasets and Metrics.} We evaluate ActWorld-JEPA on \textbf{NAVSIM v1/v2}
~\cite{navsimv1,navsimv2}, which provide multi-view observations
with map and object annotations at 2~Hz and assess planning
under non-reactive pseudo-simulation protocols.
\textbf{NAVSIM v1} reports the Predictive Driver Model Score
(PDMS), incorporating No at-fault Collision (NC), Drivable
Area Compliance (DAC), Time-to-Collision (TTC), Comfort,
and Ego Progress (EP).
\textbf{NAVSIM v2} adopts the Extended Predictive Driver Model
Score (EPDMS), additionally considering Driving Direction
Compliance (DDC), Traffic Light Compliance (TLC), Lane
Keeping (LK), History Comfort (HC), and Extended Comfort
(EC). We evaluate its navtest and more challenging navhard
splits separately.
On \textbf{nuScenes}~\cite{nuscenes}, we evaluate camera-only open-loop planning over a 6-second horizon. We report the L2 displacement error (m) to measure trajectory accuracy and the predicted Collision Rate (Col. Rate, \%) to quantify geometric collisions with surrounding agents. Both metrics are evaluated at each second from 1 to 6~s, with their average computed across the six evaluation horizons. Lower values indicate better performance.

\subsection{Implementation Details}
For NAVSIM, ActWorld-JEPA is built upon iPad~\cite{guo2026ipad}. The model uses the current multi-view observations and the preceding front-view frame to evaluate 32 candidate trajectories, each containing eight waypoints over a 4-s horizon. PC-JEPA employs eight 256-dimensional latent tokens to predict candidate-conditioned future representations, while $R^3$ exploits the resulting plan--future relations to refine the original candidate scores without altering trajectory geometry. All learnable modules are trained exclusively on NAVTRAIN using AdamW. NAVSIM v1 is trained for 20 epochs, while NAVSIM v2 adopts a two-stage future-evaluation and planning-refinement procedure. Training is conducted on NVIDIA RTX 4090 GPUs with mixed precision, and future observations are used only as training supervision and remain unavailable during inference. For nuScenes, we adopt MomAD~\cite{momad} as the baseline and follow its standard configuration with a ResNet-50~\cite{resnet} backbone, $256\times704$ input images, and six trajectory modes.

\subsection{Main Results}

\noindent\textbf{NAVSIM v1 navtest.}
Table~\ref{tab_NAVSIMv1} presents a comprehensive comparison with
end-to-end planners
\citep{liao2024diffusiondrive,diver,yao2025drivesuprim,xing2025goalflow,
shang2025drivedpo,guo2026ipad,kirby2026driving}, VLA methods
\citep{zhou2025autovla,li2025recogdrive,liu2026driveworld}, and
world-model-based approaches
\citep{zhang2026resworld,zhou2026drivedreamerpolicy,
wei2026maneuverflow,hong2026drivefuture,yang2026autojepa,
zhao2026simwam,xu2026planwam,wang2026drive}.
ActWorld-JEPA achieves the highest PDMS of 94.1 and EP of 92.2
among all compared methods, establishing leading performance
in overall planning quality and ego progress.
Notably, it surpasses the Drive-JEPA by 0.4 points
in PDMS and 0.8 points in EP, while improving DAC from
98.2 to 98.4. It also outperforms the strongest competing
method, PlanWAM, by 0.3 points in PDMS.
These results support the effectiveness of integrating
candidate-conditioned future reasoning into trajectory
selection, yielding improved planning performance
while maintaining strong drivable-area compliance.

\begin{figure*}[t]
    \centering
    \includegraphics[width=\linewidth]{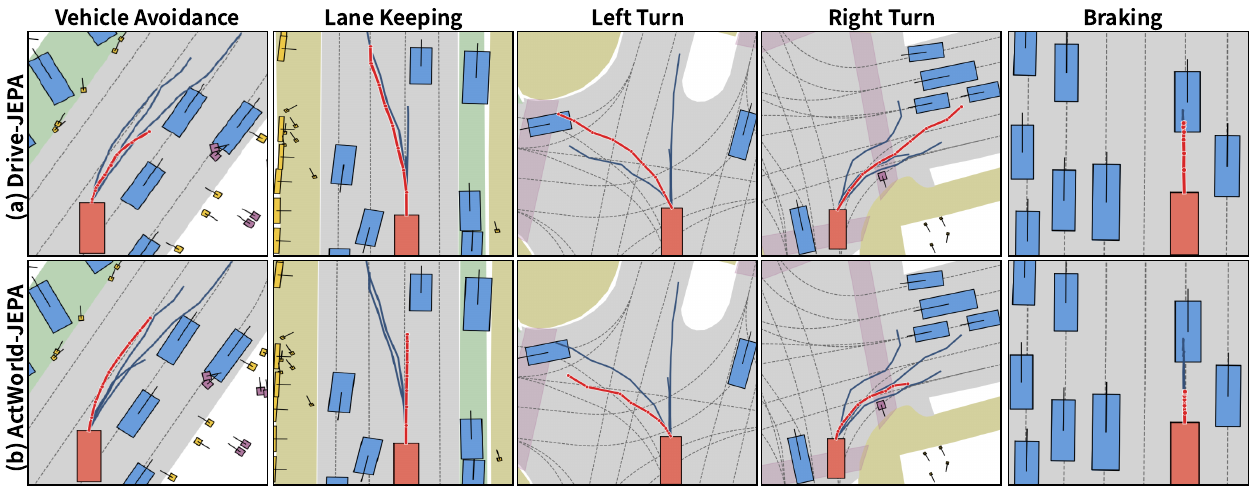}
    \caption{Qualitative comparison of trajectory selection by
    Drive-JEPA (top) and ActWorld-JEPA (bottom) across five
    challenging driving scenarios.
    Selected trajectories are shown in red.}
    \label{fig:vis}
\end{figure*}

\noindent\textbf{NAVSIM v2 navtest.}
As shown in Table~\ref{tab_navsimv2_navtest}, ActWorld-JEPA
achieves the highest EPDMS of 91.6 among the compared methods,
exceeding MomWorld~\citep{song2026momworld} by 1.5 points and
the Drive-JEPA~\citep{wang2026drive} by 3.8 points.
It also achieves the highest LK score of 97.7, with NC and TTC
scores of 98.7 and 98.2, respectively. These results support
the effectiveness of candidate-conditioned future reasoning
for trajectory selection and lane keeping.

\noindent\textbf{NAVSIM v2 navhard.}
As shown in Table~\ref{tab_navsimv2_navhard},
ActWorld-JEPA achieves 42.7 EPDMS, surpassing reproduced
Drive-JEPA~\citep{wang2026drive} by 5.0 points with improved
extended comfort in both stages. Our latent prediction avoids
pixel reconstruction, with 21.33 FPS under accelerated
deployment (Table~\ref{tab:navsim_v1_efficiency}).

\noindent\textbf{Long-Horizon Planning on nuScenes.}
Table~\ref{tab_nuscenes_planning_6s} presents the 6-second planning
results on nuScenes. ActWorld-JEPA achieves an average L2 error
of 1.20~m and a predicted collision rate of 0.76\%, outperforming
DIVER~\citep{diver} and GuideFlow~\citep{liu2025guideflow}
on both metrics. At the 6-second horizon, ActWorld-JEPA
records an L2 error of 2.39~m and a collision rate of 2.03\%.
These results demonstrate competitive long-horizon planning
performance, balancing trajectory accuracy with low predicted
collision incidence.

\subsection{Ablation Studies}
\label{sec:ablations}

\noindent\textbf{Roles of Different Components in ActWorld-JEPA.}
Table~\ref{tab:overall_planning_jepa} shows that PC-JEPA alone reaches
90.5 PDMS on NAVSIM v1 and 89.6 EPDMS on v2, improving the
reference by 3.4 and 4.3 points. Adding $R^3$ yields 94.1 PDMS
and 91.6 EPDMS, further gains of 3.6 and 2.0 points.
The full model leads across all reported metrics, supporting the
combination of candidate-conditioned future modeling and relational
reranking. Its v2 EPDMS uses corrected per-scene aggregation.

\noindent\textbf{Ablation on Core Designs in PC-JEPA.}
With $R^3$ fixed, Table~\ref{tab:pcjd_design} evaluates three core
PC-JEPA designs. Sharing futures across candidates reduces
PDMS/EPDMS from 94.1/91.6 to 92.0/89.1. Replacing recurrent
rollout with a single prediction yields 91.5/90.0, while removing
current--future interaction yields 91.8/90.3. The single-prediction
variant incurs the largest PDMS drop of 2.6 points.
Full PC-JEPA leads across all reported metrics, supporting
candidate-specific futures, recurrent prediction, and current--future
interaction as complementary components of the design.

\noindent\textbf{Ablation on Different Designs in $R^3$.}
With PC-JEPA fixed, Table~\ref{tab:fcts_design} shows that full $R^3$
achieves 94.1 PDMS/91.6 EPDMS, versus 90.5/89.6 without $R^3$.
It outperforms Simple Concat. by 1.8/0.9 points and Score Replacement
by 1.7/1.5, supporting explicit plan--future relations and residual
refinement over feature concatenation and direct score replacement.

\noindent\textbf{Model size and inference efficiency.}
Table~\ref{tab:navsim_v1_efficiency} compares official eager baselines with our
accelerated deployment on one RTX 4090.
ActWorld-JEPA achieves 21.33 FPS with 324.2M parameters,
providing $1.24\times$ the throughput of released Drive-JEPA~\cite{wang2026drive}
(17.23 FPS) for the complete model forward.
This speedup reflects the reported deployment settings.
TransFuser and DiffusionDrive remain faster but achieve lower PDMS.
All neural selector heads are counted, with shared weights deduplicated.

\subsection{Visualization}
\label{sec:visualization}

Figure~\ref{fig:vis} compares trajectory selection by
Drive-JEPA and ActWorld-JEPA across five driving scenarios.
ActWorld-JEPA selects trajectories that better accommodate
road geometry and surrounding vehicles, qualitatively
supporting the use of candidate-conditioned future reasoning
to guide trajectory selection.

\section{Conclusion}
We presented \textbf{ActWorld-JEPA}, an action-conditioned
joint-embedding world model that connects latent future
prediction with candidate-based autonomous driving.
PC-JEPA conditions recurrent latent prediction on global
and step-wise trajectory intentions, then integrates the
predicted future with the current scene through gated
cross-attention.
The resulting candidate-specific representations support
$R^3$, which models plan--future agreement and discrepancy
to produce bounded, confidence-weighted score refinements.
Together, these modules translate predicted scene evolution
into evidence for comparing candidate plans.
Benchmark-specific evaluations yield 94.1 PDMS on NAVSIM v1
navtest and 91.6 and 42.7 EPDMS on NAVSIM v2 navtest and navhard,
respectively, alongside competitive long-horizon trajectory
prediction on nuScenes. 

\noindent\textbf{Limitation and Future Work.}
ActWorld-JEPA remains dependent on the coverage of the
candidate trajectory set and the reliability of predicted
latent futures.
Future work will explore adaptive proposal generation and
uncertainty-aware candidate evaluation, alongside closed-loop
validation with reactive traffic participants.

\section*{LLM Usage Statement}
OpenAI ChatGPT and Codex were used to assist with manuscript editing,
LaTeX formatting, figure and table layout, and coding and debugging of
evaluation and inference-efficiency scripts. The authors remain responsible
for the scientific claims, experimental results, and final manuscript.

{
\small
\bibliographystyle{ieeenat_fullname}
\bibliography{main}
}

\clearpage
\normalsize
\appendix
\raggedbottom
\setcounter{table}{0}
\setcounter{figure}{0}
\setcounter{equation}{0}
\setcounter{algocf}{0}
\renewcommand{\thetable}{A\arabic{table}}
\renewcommand{\thefigure}{A\arabic{figure}}
\renewcommand{\theequation}{A\arabic{equation}}
\renewcommand{\thealgocf}{A\arabic{algocf}}
\renewcommand{\theHtable}{appendix.\arabic{table}}
\renewcommand{\theHfigure}{appendix.\arabic{figure}}
\renewcommand{\theHequation}{appendix.\arabic{equation}}
\providecommand{\theHalgocf}{}
\renewcommand{\theHalgocf}{appendix.\arabic{algocf}}
\section{Appendix}
\label{app:appendix}
This appendix complements the main paper with technical details,
benchmark-specific implementations, and additional planning results.
The material is organized as follows.
\begin{itemize}
    \setlength{\itemsep}{0pt}
    \setlength{\parskip}{0pt}
    \item Appendix~\ref{app:contributions} summarizes the contributions.
    \item Appendix~\ref{app:broader_impacts} discusses broader impacts.
    \item Appendix~\ref{app:datasets} describes the datasets and settings.
    \item Appendix~\ref{app:evaluation} defines the metrics and aggregation.
    \item Appendix~\ref{app:method} details PC-JEPA, $R^3$, and score bounds.
    \item Appendix~\ref{app:implementation} presents the implementations
    and inference settings.
    \item Appendix~\ref{app:results} reports additional planning results.
    \item Appendix~\ref{app:ablations} presents additional ablation results.
    \item Appendix~\ref{app:Qualitative} provides qualitative examples.
\end{itemize}

\subsection{Contributions}
\label{app:contributions}

Our contributions are summarized below.

\noindent\textbf{1) ActWorld-JEPA Framework.}
We propose a joint-embedding predictive framework that connects
candidate-conditioned future modeling with trajectory selection.
By relating each proposed plan to its anticipated scene evolution,
ActWorld-JEPA incorporates predictive evidence into candidate ranking.

\noindent\textbf{2) Plan-Conditioned JEPA.}
We introduce PC-JEPA to construct candidate-specific future
representations. Global trajectory intentions modulate the initial
latent state, while step-wise intentions guide recurrent prediction.
Gated current--future interaction integrates the predicted rollout
with observed context to produce future-aware planning features.

\noindent\textbf{3) Relational Residual Reranking.}
We introduce $R^3$ to translate future-aware features into candidate
preferences. Explicit plan--future agreement and discrepancy features
inform bounded, gated corrections to the base planner's scores,
connecting predictive representations to score refinement while
retaining the original scores as ranking anchors.

\noindent\textbf{4) Empirical Evaluation and Analysis.}
We evaluate ActWorld-JEPA on NAVSIM v1/v2 and nuScenes,
covering pseudo-simulation-based planning evaluation and
long-horizon open-loop planning. Comparative experiments assess
overall planning performance, while ablation studies examine
the contributions of the two modules and their key design choices,
including trajectory conditioning, latent rollout, current--future
interaction, and relational score refinement.

\subsection{Broader Impacts}
\label{app:broader_impacts}

ActWorld-JEPA may improve the reliability of autonomous driving
by incorporating anticipated action consequences into trajectory
selection. PC-JEPA integrates candidate-conditioned latent futures
with observed context, while $R^3$ translates plan--future relations
into bounded, gated refinements of candidate scores. Together,
these mechanisms may support more informed decision-making in
complex traffic by considering both proposed motions and their
predicted consequences. More broadly, this connection between
latent prediction and action evaluation may benefit model-based
planning in robotics and embodied intelligence. Realizing these
benefits in practice requires evaluation across diverse traffic
conditions and integration with system-level safety measures.

\subsection{Datasets}
\label{app:datasets}

\noindent\textbf{NAVSIM.}
We evaluate ActWorld-JEPA on NAVSIM v1 \textit{navtest},
NAVSIM v2 \textit{navtest}, and NAVSIM v2
\textit{navhard}~\cite{navsimv1,navsimv2}.
These settings assess ego-trajectory quality using recorded
driving scenes and structured driving criteria.
NAVSIM v1 evaluates planned trajectories against recorded,
non-reactive background traffic using the Predictive Driver
Model Score (PDMS).
For NAVSIM v2, we report the Extended Predictive Driver Model
Score (EPDMS), distinguishing the per-scene average on
\textit{navtest} from the two-stage evaluation on \textit{navhard}.
The latter combines an initial real scene with precomputed
synthetic follow-up observations, weighted according to their
proximity to the first-stage endpoint.
This setting assesses planning under deviations from the
recorded trajectory.
Metric definitions and aggregation procedures are detailed
in Appendix~\ref{app:evaluation}.

\noindent\textbf{nuScenes.}
We conduct camera-only open-loop planning experiments on
nuScenes~\cite{nuscenes}, following the training and validation
setting described in Section~\ref{experiments}.
The dataset comprises 1,000 real-world driving scenes of
approximately 20 seconds each, recorded with a surround-view
sensor suite.
Under the open-loop protocol, predicted ego trajectories are
evaluated against logged data without altering subsequent
traffic evolution.
We use a single 6-s planning setting, predicting twelve
waypoints at 2~Hz. All reported horizons are evaluated
from these 6-s predictions.
The long-horizon evaluation reports L2 displacement error
and predicted collision rate at each second from 1 to 6~s,
complemented by Trajectory Prediction Consistency (TPC)
over 4--6~s.
These evaluations characterize trajectory accuracy, predicted
collision incidence, and temporal consistency across planning
horizons.

\subsection{Evaluation Metrics}
\label{app:evaluation}

We evaluate planning quality using PDMS and EPDMS on
NAVSIM v1/v2, and evaluate open-loop planning on nuScenes
using displacement error, predicted collision rate,
and trajectory prediction consistency.

\noindent\textbf{NAVSIM v1.}
The Predictive Driver Model Score (PDMS) evaluates ego
trajectories against non-reactive background
traffic~\cite{navsimv1}.
It combines multiplicative compliance penalties with
weighted driving-quality criteria:
\begin{equation}
s_n^{\mathrm{v1}}
=
m_{n,\mathrm{NC}}m_{n,\mathrm{DAC}}
\frac{
5m_{n,\mathrm{EP}}
+5m_{n,\mathrm{TTC}}
+2m_{n,\mathrm C}
}{12},
\label{eq:app_pdms}
\end{equation}
where $m_{n,q}\in[0,1]$ denotes component $q$ for scene $n$.
No at-fault collision (NC) penalizes collisions attributable
to the ego vehicle, while drivable-area compliance (DAC)
assesses whether the vehicle remains within the drivable
region. Ego progress (EP) measures advancement along the
route, time-to-collision (TTC) evaluates the prescribed
collision-time margin, and comfort (C) assesses motion
smoothness. The multiplicative terms penalize compliance
violations, while the weighted terms balance progress,
collision avoidance, and comfort.

\noindent\textbf{NAVSIM v2.}
The Extended Predictive Driver Model Score (EPDMS)
introduces additional rule-compliance and comfort
criteria~\cite{navsimv2}.
Its multiplicative and weighted component sets are
\begin{align}
\mathcal M_2
&=\{\mathrm{NC},\mathrm{DAC},\mathrm{DDC},\mathrm{TLC}\},
\nonumber\\
\mathcal W_2
&=\{\mathrm{EP},\mathrm{TTC},\mathrm{LK},\mathrm{HC},\mathrm{EC}\}.
\label{eq:app_epdms_components}
\end{align}
Driving-direction compliance (DDC) and traffic-light
compliance (TLC, abbreviated as TL in the tables) assess
adherence to the corresponding traffic rules.
Lane keeping (LK) evaluates lane-following behavior.
History comfort (HC) incorporates preceding ego motion
into comfort evaluation, whereas extended comfort (EC)
assesses motion consistency across successive planning
outputs.

Human-reference filtering suppresses eligible penalties
when the corresponding violation is also present in the
recorded human trajectory.
Let $m^{\mathrm{agent}}_{n,q}$ and
$m^{\mathrm{human}}_{n,q}$ denote the agent and reference
scores, respectively, and let $a_{n,q}$ indicate whether
filtering is applicable and enabled.
The filtered component is
\begin{equation}
\bar m_{n,q}
=
\begin{cases}
1,
& a_{n,q}=1
  \ \text{and}\ m^{\mathrm{human}}_{n,q}=0,\\
m^{\mathrm{agent}}_{n,q},
& \text{otherwise}.
\end{cases}
\label{eq:app_human_filter}
\end{equation}
Filtering is applied to individual components before
recomputing the composite score.

Let $\mathcal W_{2,n}\subseteq\mathcal W_2$ contain the
weighted components available for scene $n$.
The per-scene EPDMS is
\begin{equation}
e_n
=
\left(
\prod_{q\in\mathcal M_2}\bar m_{n,q}
\right)
\frac{
\sum_{q\in\mathcal W_{2,n}}w_q\bar m_{n,q}
}{
\sum_{q\in\mathcal W_{2,n}}w_q
},
\label{eq:app_epdms}
\end{equation}
with weights
\begin{equation}
(w_{\mathrm{EP}},w_{\mathrm{TTC}},w_{\mathrm{LK}},
 w_{\mathrm{HC}},w_{\mathrm{EC}})
=(5,5,2,2,2).
\label{eq:app_epdms_weights}
\end{equation}
Unavailable components are omitted from both sums.
In particular, when EC is unavailable, the denominator
decreases from 16 to 14 rather than treating the missing
component as zero.
Component thresholds and filtering eligibility follow
the NAVSIM devkit~\cite{navsimdevkitmetrics}.

\noindent\textbf{Navtest Aggregation.}
For the evaluated scene set $\mathcal S$, the reported
NAVSIM v1 and v2 \textit{navtest} scores are
\begin{equation}
\begin{aligned}
\operatorname{PDMS}_{\mathrm{navtest}}
&=
\frac{100}{|\mathcal S|}
\sum_{n\in\mathcal S}s_n^{\mathrm{v1}},\\
\operatorname{EPDMS}_{\mathrm{navtest}}
&=
\frac{100}{|\mathcal S|}
\sum_{n\in\mathcal S}e_n.
\end{aligned}
\label{eq:app_navtest_aggregation}
\end{equation}
Each component is averaged over the scenes for which
it is available. Composite scores are computed before
dataset-level averaging and therefore cannot generally
be reconstructed from the component means reported
in the tables.

\noindent\textbf{Navhard Two-Stage Aggregation.}
NAVSIM v2 \textit{navhard} evaluates initial observations
together with precomputed synthetic follow-up scenes.
For an initial evaluation branch $b$, let
$\hat{\mathbf x}_b$ denote the first-stage endpoint
and $\mathbf x_{b,j}$ the starting position of follow-up
scene $j\in\mathcal F_b$.
The proximity weights are
\begin{equation}
\begin{aligned}
u_{b,j}
&=
\exp\left(
-\frac{
\|\mathbf x_{b,j}-\hat{\mathbf x}_b\|_2^2
}{2\sigma^2}
\right),\\
\omega_{b,j}
&=
\frac{u_{b,j}}
{\sum_{\ell\in\mathcal F_b}u_{b,\ell}},
\end{aligned}
\label{eq:app_navhard_weights}
\end{equation}
where $\sigma^2$ is the kernel parameter specified by
the evaluator.
Uniform weights are used when the normalization sum
is numerically zero or undefined.

Given first-stage EPDMS $e_b^{(1)}$ and follow-up scores
$e_{b,j}^{(2)}$, the branch score is
\begin{equation}
c_b
=
e_b^{(1)}
\sum_{j\in\mathcal F_b}
\omega_{b,j}e_{b,j}^{(2)}.
\label{eq:app_navhard_branch}
\end{equation}
The official evaluation mapping pairs the current and
preceding initial frames.
For $G$ valid mapping groups, the final score is
\begin{equation}
\operatorname{EPDMS}_{\mathrm{navhard}}
=
\frac{100}{G}
\sum_{g=1}^{G}
\frac{c_{g,0}+c_{g,1}}{2},
\label{eq:app_navhard_aggregation}
\end{equation}
where $c_{g,0}$ and $c_{g,1}$ denote the branch scores
associated with the two initial frames in group $g$.
Thus, stage scores are multiplied within each branch
before averaging across groups.
All reported NAVSIM scores use a 0--100 scale, with
higher values indicating better performance.

\noindent\textbf{nuScenes.}
Following the MomAD evaluator~\citep{momad}, we report
L2 displacement error, predicted collision rate, and
Trajectory Prediction Consistency (TPC).
Let $\mathcal V$ denote samples with complete 6-s future
annotations and $K_T=2T$ the number of prediction steps
up to horizon $T$ seconds.
The metrics are computed as
\begin{align}
\operatorname{L2}(T)
&=
\frac{1}{|\mathcal V|K_T}
\sum_{n\in\mathcal V}\sum_{k=1}^{K_T}
\left\|
\hat{\mathbf p}_{n,k}
-\mathbf p^{\mathrm{ref}}_{n,k}
\right\|_2,
\label{eq:app_nuscenes_l2}\\
\operatorname{Col}(T)
&=
\frac{100}{|\mathcal V|K_T}
\sum_{n\in\mathcal V}\sum_{k=1}^{K_T}
\hat c_{n,k}(1-c^{\mathrm{ref}}_{n,k}),
\label{eq:app_nuscenes_collision}\\
\operatorname{TPC}(T)
&=
\frac{1}{|\mathcal V|K_T}
\sum_{n\in\mathcal V}\sum_{k=1}^{K_T}
\left\|
\hat{\mathbf p}_{n,k}-\mathbf h_{n,k}
\right\|_2.
\label{eq:app_tpc}
\end{align}
Here, $\hat{\mathbf p}_{n,k}$ and
$\mathbf p^{\mathrm{ref}}_{n,k}$ are the predicted and
reference ego positions, while $\mathbf h_{n,k}$ is the
cached historical prediction under the MomAD evaluator's
coordinate and waypoint-index conventions.
The indicators $\hat c_{n,k}$ and $c^{\mathrm{ref}}_{n,k}$
identify intersections between surrounding actors and
the predicted or reference ego footprint, respectively.
Reference collisions are masked from the numerator
without changing the denominator.

L2 and collision rate are reported at 1--6~s, with
their \textit{Avg.} entries computed as
\begin{equation}
\operatorname{Avg}(M)
=
\frac{1}{6}\sum_{j=1}^{6}M(j),
\qquad
M\in\{\operatorname{L2},\operatorname{Col}\},
\label{eq:app_nuscenes_average}
\end{equation}
where $j$ denotes the horizon in seconds.
TPC is reported at 4--6~s without an additional
cross-horizon average.
L2 and TPC are measured in meters, whereas collision rate
is reported as a percentage. Lower values are preferable for
all three metrics.

\subsection{Additional Details of ActWorld-JEPA}
\label{app:method}

\noindent\textbf{Organization of Candidate-Conditioned Future Evidence.}
PC-JEPA uses shared prediction parameters to construct
a latent future for each candidate trajectory.
The candidate index is preserved through rollout,
current--future interaction, and relational scoring,
ensuring that each predicted future contributes to
the preference for the plan that conditions it.

For $L$ latent tokens of width $D$ at each prediction
step, the rollout is organized into the sequence
\begin{equation}
\mathbf F_t^k
=
\operatorname{Concat}_{\mathrm{tokens}}
\left(
\widehat{\mathbf Z}_{t+1}^{\,k},
\ldots,
\widehat{\mathbf Z}_{t+H}^{\,k}
\right),
\label{eq:app_future_tokens}
\end{equation}
where $\mathbf F_t^k\in\mathbb R^{HL\times D}$.
This representation retains the complete rollout
rather than only the terminal state.
For current query token $i$, the attended future
feature within one attention head is
\begin{equation}
\mathbf u_{t,i}^k
=
\sum_{h=1}^{H}\sum_{\ell=1}^{L}
a_{t,i,h,\ell}^k
\left(
\widehat{\mathbf z}_{t+h,\ell}^{\,k}\mathbf W_V
\right),
\label{eq:app_future_evidence}
\end{equation}
where the attention coefficients are nonnegative
and sum to one over the future tokens. Multi-head attention
combines the head-specific outputs through an output projection.
Recurrent prediction encodes the evolution associated
with the candidate, while attention determines which
parts of this evolution inform the current representation.
The interaction therefore aggregates evidence across
prediction steps and latent locations without imposing
a fixed correspondence between current and future tokens.

\noindent\textbf{Complementary Roles of Feature and Score Gating.}
The two gates regulate different stages of the
prediction--selection pathway.
Writing $\mathbf U_t^k$ for the attended future
features after combining all attention heads, the PC-JEPA
representation is
\begin{equation}
\mathbf H_t^k
=
\operatorname{LN}
\left(
\mathbf Z_t+
\boldsymbol\alpha_t^k\odot\mathbf U_t^k
\right).
\label{eq:app_feature_gate}
\end{equation}
The feature gate controls the contribution of predicted
evidence to the current representation.
As this gate approaches zero, the representation
approaches the normalized current-scene features.
This does not necessarily recover the baseline score,
since the relation head still receives the candidate
intention and current-scene information.
By contrast, as the score gate in $R^3$ approaches zero,
the residual correction vanishes.

The agreement and discrepancy features expose
interactions between the projected plan and future
representations; they do not impose a fixed ordering
based on feature similarity.
Their contribution to candidate preference is learned
through the selection objective.
For $\mathcal L_{\mathrm{sel}}=-\log\pi_t^{k^+}$,
\begin{equation}
\frac{\partial\mathcal L_{\mathrm{sel}}}
     {\partial\widetilde s_t^k}
=
\frac{\pi_t^k-\mathbb I[k=k^+]}{\tau},
\label{eq:app_selection_gradient}
\end{equation}
where $\tau>0$ denotes the Softmax temperature and
$k^+$ is treated as a fixed supervision label.
This gradient trains the correction and its gate
through their effect on selection.
When the predictive pathway is trainable, the signal
also propagates through relational scoring and
current--future interaction into PC-JEPA.
The score gate is thus a learned modulation coefficient,
rather than a separately calibrated probability of
prediction correctness.

\noindent\textbf{Association of Future and Selection Supervision.}
The prediction and selection objectives use different
associations with the recorded trajectory.
Let $\tau_t^+$ denote the recorded plan corresponding
to the observed future, and let $\mathcal R_\theta$
denote the PC-JEPA rollout operator.
The branch supervised by the future loss is
\begin{equation}
\widehat{\mathcal Z}_t^+
=
\mathcal R_\theta(\mathbf Z_t,\tau_t^+).
\label{eq:app_recorded_rollout}
\end{equation}
The superscript $+$ identifies the recorded-plan branch,
not an index in the generated candidate set.
Its predicted states are matched to the detached
future embeddings in Eq.~\eqref{eq:future_loss}.
The recorded plan can therefore condition prediction
supervision without being added to the proposals
available for selection.

Selection supervision instead identifies the candidate
closest to the expert trajectory:
\begin{equation}
k^+
\in
\operatorname*{arg\,min}_{k\in\{1,\ldots,K\}}
\frac{1}{H}
\sum_{h=1}^{H}
\left\|
\mathbf p_{t+h}^k-
\mathbf p_{t+h}^{\mathrm{expert}}
\right\|_2.
\label{eq:app_selection_target}
\end{equation}
Here, $\mathbf p_{t+h}^{\mathrm{expert}}$ is the position
at step $h$ of the recorded plan $\tau_t^+$. Ties are resolved
using a fixed rule. The closest candidate need not coincide
with the recorded plan.
Its label specifies a preference within the proposal
set, whereas the observed future provides a target
for prediction under the recorded plan.
Keeping these associations distinct avoids assigning
the same observed outcome indiscriminately to
alternative candidate trajectories.

Algorithm~\ref{alg:app_prediction_selection} connects
the two supervision branches to the shared scoring
pathway.
It summarizes the formulation in Section~\ref{method},
using PC-JEPA and the relational head as defined there.
The recorded-plan rollout and candidate rollouts share
the predictive operator, but only the former receives
the observed-future matching target.
Candidate selection is trained through the score
distribution before the inference-time argmax.
The baseline integration and architecture settings are
summarized in Appendix~\ref{app:implementation}.

\begin{algorithm}[t]
\caption{Prediction--Selection Coupling in ActWorld-JEPA}
\label{alg:app_prediction_selection}
\footnotesize
\DontPrintSemicolon

\KwIn{
Available observations $\mathcal O_{\leq t}$;
mode $\mu\in\{\mathrm{train},\mathrm{infer}\}$;
residual strength $\lambda\geq0$;
temperature $\tau>0$; loss weights $\lambda_f,\lambda_s\geq0$.
Training additionally provides the recorded plan
$\tau_t^+$ and corresponding future observations.
}
\KwOut{
Training objective $\mathcal L$ in training mode;
selected trajectory $\hat\tau_t$ in inference mode.
}

$\mathbf Z_t\leftarrow E(\mathcal O_{\leq t})$\;
$\{(\tau_t^k,b_t^k)\}_{k=1}^{K}
\leftarrow\operatorname{BasePlanner}(\mathbf Z_t)$\;

\For{$k\leftarrow1$ \KwTo $K$}{
    $(\mathbf C_t^k,\mathbf H_t^k)
    \leftarrow\operatorname{PC\text{-}JEPA}
    (\mathbf Z_t,\tau_t^k)$\;
    $(\Delta s_t^k,\rho_t^k)
    \leftarrow\operatorname{RelationHead}
    (\mathbf C_t^k,\mathbf H_t^k)$\;
    $\widetilde s_t^k
    \leftarrow b_t^k+
    \lambda\sigma(\rho_t^k)\tanh(\Delta s_t^k)$\;
}
$\boldsymbol\pi_t
\leftarrow
\operatorname{Softmax}
(\widetilde{\mathbf s}_t/\tau)$\;

\eIf{$\mu=\mathrm{train}$}{
    $k^+\leftarrow$ candidate index obtained from
    Eq.~\eqref{eq:app_selection_target}\;
    $\mathcal L_{\mathrm{sel}}
    \leftarrow-\log\pi_t^{k^+}$\;

    $\widehat{\mathcal Z}_t^+
    \leftarrow\mathcal R_\theta(\mathbf Z_t,\tau_t^+)$\;
    \For{$h\leftarrow1$ \KwTo $H$}{
        $\mathbf Z_{t+h}^+\leftarrow$
        embedding of the corresponding observed future\;
        $\mathbf Y_{t+h}
        \leftarrow\operatorname{sg}(\mathbf Z_{t+h}^+)$\;
    }
    $\mathcal L_{\mathrm{future}}
    \leftarrow
    \frac{1}{H}\sum_{h=1}^{H}
    \|\widehat{\mathbf Z}_{t+h}^{\,+}
    -\mathbf Y_{t+h}\|_F^2$\;
    $\mathcal L_{\mathrm{plan}}\leftarrow$
    the base planner's training objective\;
    $\mathcal L
    \leftarrow
    \mathcal L_{\mathrm{plan}}
    +\lambda_f\mathcal L_{\mathrm{future}}
    +\lambda_s\mathcal L_{\mathrm{sel}}$\;
    \Return{$\mathcal L$}\;
}{
    $k^\star\leftarrow
    \operatorname*{arg\,max}_{k}\pi_t^k$\;
    $\hat\tau_t\leftarrow\tau_t^{k^\star}$\;
    \Return{$\hat\tau_t$}\;
}
\end{algorithm}

Here, $\|\cdot\|_F$ denotes the Frobenius norm over
the latent-token matrix.
Stop-gradient blocks the future loss from updating
the target embeddings through the target branch.
It does not itself freeze an encoder shared with
other trainable branches.
The encoder update policy is therefore a separate
optimization choice.
At inference, neither the recorded plan nor future
observations enter the scoring pathway, and the
reranking operation selects an existing proposal
without changing its geometry.

\begin{table}[t]
\centering
\caption{
Trajectory prediction consistency on the nuScenes validation
set. TPC is reported in meters at 4--6~s, with lower values
indicating better temporal consistency.
}
\renewcommand{\arraystretch}{0.75}
\setlength{\tabcolsep}{4.6mm}
\resizebox{\linewidth}{!}{
\begin{tabular}{l c c c c}
\toprule
& & \multicolumn{3}{c}{TPC (m)$\downarrow$} \\
\cmidrule(lr){3-5}
Method & Venue & 4s & 5s & 6s \\
\midrule

UniAD~\cite{uniad}
& CVPR'23 & 1.49 & 1.81 & 2.41 \\

VAD~\cite{jiang2023vad}
& ICCV'23 & 1.55 & 1.73 & 2.17 \\

SparseDrive~\cite{sun2024sparsedrive}
& ICRA'25 & 1.33 & 1.66 & 1.99 \\

MomAD~\cite{momad}
& CVPR'25 & 1.19 & 1.45 & 1.61 \\

\rowcolor{red!5}
ActWorld-JEPA (Ours)
& -- & \textbf{1.04} & \textbf{1.31} & \textbf{1.60} \\

\bottomrule
\end{tabular}}
\label{tab:tpc_nuscenes}
\end{table}
\begin{table}[t]
\centering
\caption{
Ablation study of the two core modules on the nuScenes validation set. Results are reported at the challenging 4--6\,s horizons. Lower is better for all metrics.
}
\renewcommand{\arraystretch}{0.75}
\setlength{\tabcolsep}{3.8mm}
\resizebox{\linewidth}{!}{
\begin{tabular}{c c c c c c c c}
\toprule
\multicolumn{2}{c}{Components}
& \multicolumn{3}{c}{L2 (m)$\downarrow$}
& \multicolumn{3}{c}{Col. Rate (\%)$\downarrow$} \\
\cmidrule(lr){1-2}
\cmidrule(lr){3-5}
\cmidrule(lr){6-8}
PC-JEPA & $R^3$
& 4\,s & 5\,s & 6\,s
& 4\,s & 5\,s & 6\,s \\
\midrule

&
& 1.67 & 1.98 & 2.45
& 0.83 & 1.43 & 2.13 \\

\checkmark &
& 1.32 & 1.83 & 2.41
& 0.75 & 1.38 & 2.09 \\

\rowcolor{red!5}
\checkmark & \checkmark
& \textbf{1.31} & \textbf{1.82} & \textbf{2.39}
& \textbf{0.72} & \textbf{1.33} & \textbf{2.03} \\

\bottomrule
\end{tabular}}
\label{tab:nuscenes_overall_ablation}
\end{table}

\begin{table}[t]
\centering
\caption{
Detailed NAVSIM v2 navtest results for the core PC-JEPA variants in Table~\ref{tab:pcjd_design}, with $R^3$ fixed. Baseline and Full follow Table~\ref{tab:overall_planning_jepa}. Other variant subscores are from the detailed ablation records; EPDMS follows Table~\ref{tab:pcjd_design}. C--F denotes current--future interaction. Bold indicates the best result at the displayed precision.
}
\renewcommand{\arraystretch}{0.75}
\setlength{\tabcolsep}{0.6mm}
\resizebox{\linewidth}{!}{
\begin{tabular}{l c c c c c c c c c c}
\toprule
Variant
& NC$\uparrow$ & DAC$\uparrow$ & DDC$\uparrow$ & TLC$\uparrow$
& EP$\uparrow$ & TTC$\uparrow$ & LK$\uparrow$ & HC$\uparrow$
& EC$\uparrow$ & EPDMS$\uparrow$ \\
\midrule

Baseline
& 96.9 & 98.3 & 97.9 & 97.6
& 87.1 & 97.1 & 96.1 & 95.2
& 84.2 & 85.3 \\
\midrule

Shared Future
& 97.9 & 98.5 & 98.7 & 98.5
& 88.3 & 97.2 & 97.0 & 97.0
& 83.2 & 89.1 \\

Single-Step Prediction
& 98.5 & 98.6 & 99.0 & 99.8
& 88.6 & 98.1 & 97.6 & 97.1
& 83.6 & 90.0 \\

w/o C--F Interaction
& 98.4 & 98.6 & \textbf{99.1} & 99.8
& 88.6 & 97.9 & 97.6 & 97.4
& 83.4 & 90.3 \\

\rowcolor{red!5}
Full PC-JEPA
& \textbf{98.7} & \textbf{98.7} & \textbf{99.1} & \textbf{99.9}
& \textbf{88.7} & \textbf{98.2} & \textbf{97.7} & 97.9
& 85.0 & \textbf{91.6} \\

\bottomrule
\end{tabular}}
\label{tab:pcjepa_v2_detailed}
\end{table}

\begin{table}[t]
\centering
\caption{
Detailed NAVSIM v2 navtest results for the $R^3$ variants in Table~\ref{tab:fcts_design}, with PC-JEPA fixed. Baseline, w/o $R^3$, and Full follow Table~\ref{tab:overall_planning_jepa}. Other variant subscores are from the detailed ablation records; EPDMS follows Table~\ref{tab:fcts_design}. Bold indicates the best result at the displayed precision.
}
\renewcommand{\arraystretch}{0.75}
\setlength{\tabcolsep}{0.6mm}
\resizebox{\linewidth}{!}{
\begin{tabular}{l c c c c c c c c c c}
\toprule
Variant
& NC$\uparrow$ & DAC$\uparrow$ & DDC$\uparrow$ & TLC$\uparrow$
& EP$\uparrow$ & TTC$\uparrow$ & LK$\uparrow$ & HC$\uparrow$
& EC$\uparrow$ & EPDMS$\uparrow$ \\
\midrule

Baseline
& 96.9 & 98.3 & 97.9 & 97.6
& 87.1 & 97.1 & 96.1 & 95.2
& 84.2 & 85.3 \\
\midrule

\multicolumn{11}{c}{\textit{Plan--Future Relations}} \\
\midrule

w/o $R^3$
& 98.4 & 98.6 & 98.8 & 99.6
& 88.1 & 98.0 & 97.2 & 97.6
& 84.8 & 89.6 \\

Plan-Only Features
& 98.5 & 98.6 & \textbf{99.1} & 99.8
& 88.6 & 97.9 & 97.7 & 97.9
& 84.9 & 90.5 \\

Simple Concat.
& 98.6 & \textbf{98.7} & \textbf{99.1} & 98.7
& \textbf{88.7} & 98.1 & 97.7 & 97.9
& 84.7 & 90.7 \\

w/o Compatibility
& 98.6 & 98.6 & \textbf{99.1} & 99.2
& 88.5 & 98.1 & 97.7 & 97.9
& \textbf{85.0} & 90.0 \\

\midrule
\multicolumn{11}{c}{\textit{Score Refinement}} \\
\midrule

Score Replacement
& 98.3 & 97.7 & 98.8 & 99.7
& 88.0 & 98.0 & 97.1 & 97.6
& 84.9 & 90.1 \\

\rowcolor{red!5}
Full $R^3$
& \textbf{98.7} & \textbf{98.7} & \textbf{99.1} & \textbf{99.9}
& \textbf{88.7} & \textbf{98.2} & 97.7 & 97.9
& \textbf{85.0} & \textbf{91.6} \\

\bottomrule
\end{tabular}}
\label{tab:r3_v2_detailed}
\end{table}

\noindent\textbf{Residual Bounds and Ranking Preservation.}
The bounded residual limits how strongly predictive
evidence can modify the baseline preference.
At a fixed planning step, omit the time subscript
and let $c_k=\sigma(\rho_k)$.
The correction is
\begin{equation}
\delta_k
=
\lambda c_k\tanh(\Delta s_k).
\label{eq:app_residual_definition}
\end{equation}
Its magnitude is bounded by $\lambda c_k$ and hence
by $\lambda$.
For candidates $i$ and $j$, the change in their score
difference satisfies
\begin{equation}
\left|
(\widetilde s_i-\widetilde s_j)
-
(b_i-b_j)
\right|
\leq
\lambda(c_i+c_j).
\label{eq:app_pair_bound}
\end{equation}
This follows from the triangle inequality applied
to the two residual corrections.
A sufficient condition for preserving the baseline
preference for candidate $i$ over candidate $j$ is
\begin{equation}
b_i-b_j
>
\lambda(c_i+c_j).
\label{eq:app_order_preservation}
\end{equation}
In particular, a baseline margin greater than
$2\lambda$ cannot be reversed.
If this condition holds between the unique baseline
maximizer and every competing candidate, the selected
trajectory remains unchanged.
Smaller margins permit, but do not require, a change
in ranking.

These bounds explain how baseline anchoring and
future-dependent refinement coexist: sufficiently
large baseline margins are preserved, while close
preferences remain adjustable.
Setting $\lambda=0$ recovers the baseline scores
and, under the same tie-breaking rule, the baseline
selection.
For fixed scores, a positive Softmax temperature
changes probability concentration but not the
maximizing indices, although it affects the training
gradients in Eq.~\eqref{eq:app_selection_gradient}.
These guarantees apply to the bounded $R^3$ scores and
to selection by their argmax. Additional acceptance rules,
if used, require separate analysis.
\subsection{Implementation Details}
\label{app:implementation}

\noindent\textbf{Architecture and Trajectory Settings.}
We build ActWorld-JEPA upon iPad~\citep{guo2026ipad}
for NAVSIM and MomAD~\citep{momad} for nuScenes.
The base planner provides scene representations, candidate
trajectories, and initial candidate scores. The NAVSIM
configuration uses a ViT-L visual backbone and 32 candidate
trajectories over a $4\,\mathrm{s}$ planning horizon, with eight
waypoints per candidate at $0.5\,\mathrm{s}$ intervals, eight
scene-pooling queries, a latent feature dimension of 256, and
eight PC-JEPA attention heads. The nuScenes configuration
uses a ResNet-50 backbone and six candidates over a
$6\,\mathrm{s}$ horizon, with 12 waypoints per candidate at the
same interval, one scene-pooling query, a latent feature
dimension of 128, and four PC-JEPA attention heads.
PC-JEPA introduces candidate-conditioned latent future
prediction, while $R^3$ incorporates plan--future relations
into candidate ranking.

\noindent\textbf{PC-JEPA Configuration.}
The candidate waypoints define the temporal grid for
trajectory conditioning and recurrent latent prediction.
Global and step-wise intentions condition the rollout
as described in the main text, and gated current--future
interaction integrates the predicted sequence with
the observed context. The NAVSIM attention modules
use a dropout rate of 0.05.

\noindent\textbf{$R^3$ Integration.}
Each candidate-plan feature is paired with its
corresponding future-aware representation throughout
relational scoring.
The resulting agreement and discrepancy features
are mapped to bounded, gated corrections to the
original candidate scores.
The corrected preferences are then used for trajectory
selection, with the baseline scores providing the
reference for residual adjustment.

\noindent\textbf{NAVSIM Setup.}
We initialize the NAVSIM models from the corresponding
pretrained iPad checkpoints.
Each planning step evaluates 32 candidate trajectories
over a 4-s horizon, represented by eight waypoints
at 0.5-s intervals.
The same temporal resolution is used for
candidate-conditioned latent rollout.
The learnable components are fitted on NAVTRAIN.
Evaluation covers NAVSIM v1 navtest, NAVSIM v2 navtest,
and NAVSIM v2 navhard, with their scoring and aggregation
procedures specified in Appendix~\ref{app:evaluation}.
For the reported NAVSIM v1 configuration, the final
global selection coefficients and acceptance threshold
are calibrated using aggregate NAVTEST feedback.

\noindent\textbf{nuScenes Setup.}
For nuScenes, we initialize ActWorld-JEPA from the
pretrained camera-only MomAD baseline and integrate
PC-JEPA and $R^3$ into its planning branch.
The baseline scene representations and trajectory
candidates provide the inputs for adapting the added
future-modeling and reranking modules.

We retain the ResNet-50 backbone, an input resolution
of $256\times704$, and six trajectory modes.
The model predicts twelve waypoints over a 6-s horizon
at 2\,Hz.
All reported temporal horizons are evaluated from
these 6-s predictions using the MomAD evaluation
protocol; no separate 3-s model is trained.

\noindent\textbf{Inference Procedure.}
Online planning uses the available scene observations,
ego state, and candidate trajectories supplied by
the base planner.
PC-JEPA predicts latent futures conditioned on these
candidates, and $R^3$ uses the resulting relational
features to adjust their preferences.
Future observations, reference trajectories, and
ground-truth outcome labels used for supervision
are unavailable during inference.
The additional computation associated with the two
modules arises from latent rollout, current--future
interaction, and relational candidate scoring.

\begin{figure*}[t]
\centering
\includegraphics[width=\linewidth]{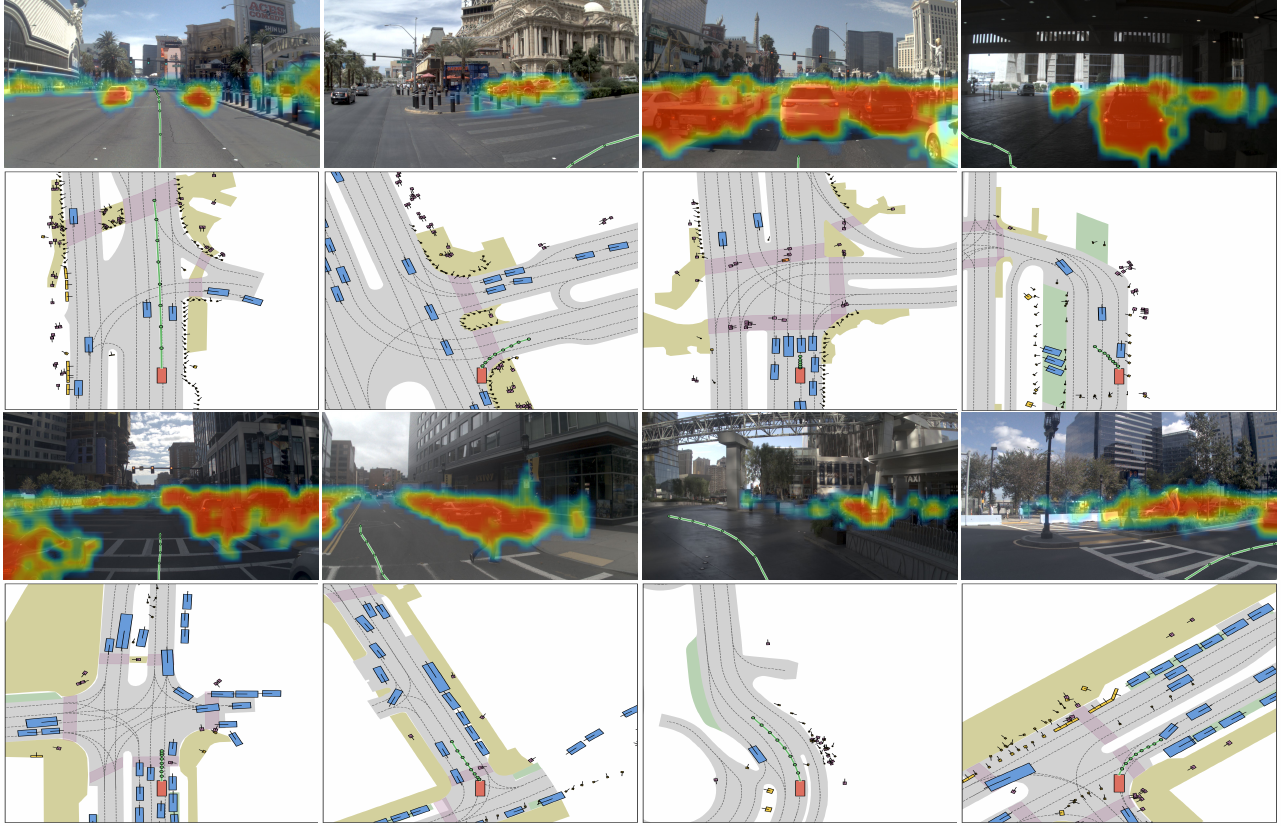}
\caption{\textbf{Additional qualitative planning results on NAVSIM.}
Front-view semantic overlays and corresponding bird's-eye-view trajectories
across diverse driving scenarios illustrate the correspondence between
semantic scene awareness and planning decisions. ActWorld-JEPA generates
plausible trajectories adapted to road geometry and surrounding traffic,
illustrating scene-dependent trajectory selection.}
\label{vis_planning_nav}
\end{figure*}

\begin{figure*}[t]
\centering
\includegraphics[width=\linewidth]{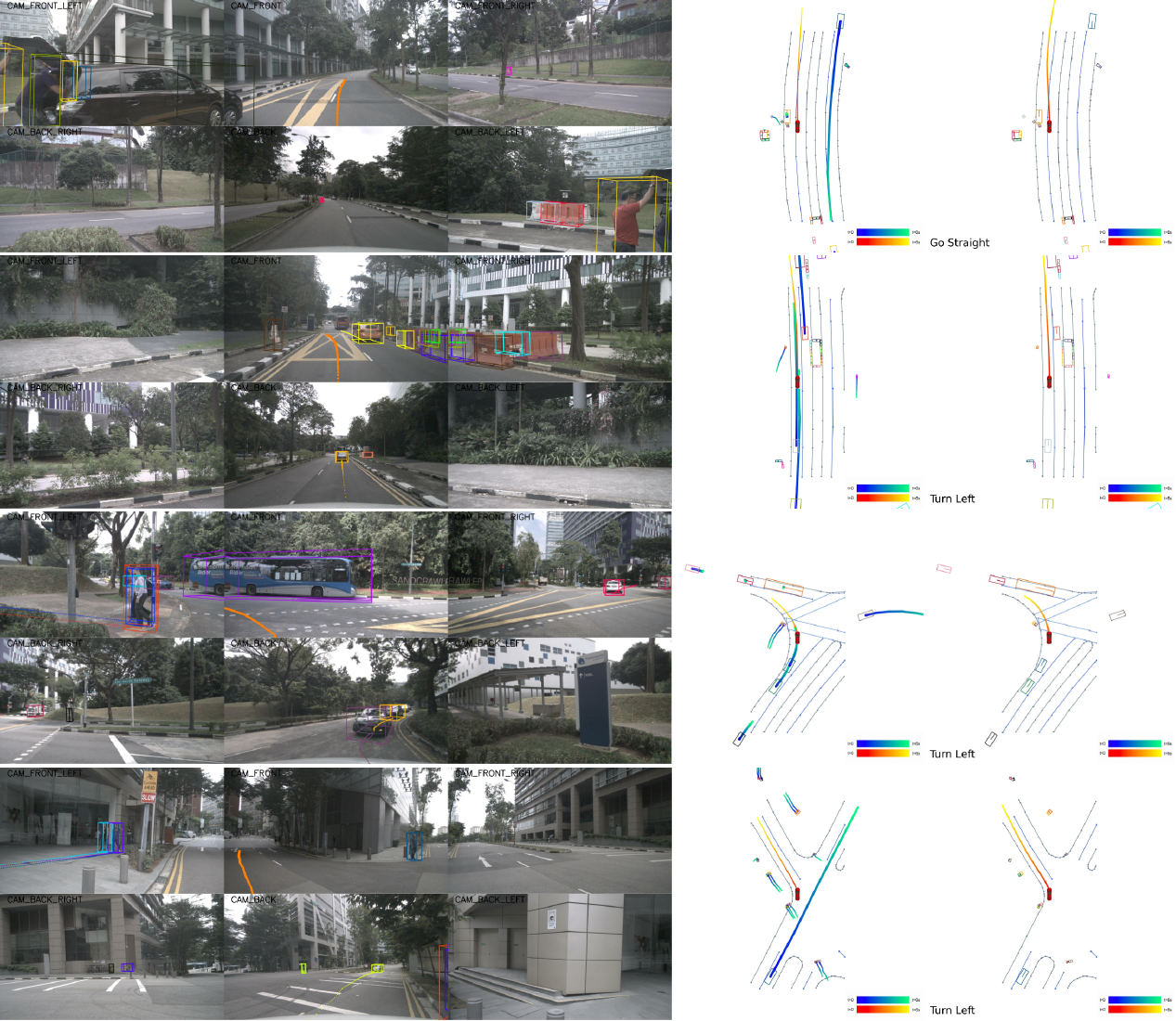}
\caption{\textbf{Qualitative 6-second planning results on nuScenes.}
ActWorld-JEPA generates coherent long-horizon trajectories across diverse
driving scenarios, adapting to road geometry and surrounding traffic while
producing plausible planned motions.}
\label{nus_vis_fl}
\end{figure*}

\subsection{More Planning Results}
\label{app:results}

\noindent\textbf{Trajectory Prediction Consistency.}
Table~\ref{tab:tpc_nuscenes} evaluates temporal planning
consistency on the nuScenes validation set.
ActWorld-JEPA achieves the lowest TPC among the compared
methods at all three horizons, with values of 1.04,
1.31, and 1.60~m at 4, 5, and 6~s, respectively.
Compared with the MomAD baseline, these correspond
to reductions of 0.15, 0.14, and 0.01~m.
The improvements are more pronounced at 4 and 5~s,
while performance remains comparable at 6~s.
These results complement the displacement and collision
metrics by showing lower prediction discrepancy under
the same TPC evaluation convention.

\subsection{More Ablation Results}
\label{app:ablations}

\noindent\textbf{Roles of Different Components on nuScenes.}
Table~\ref{tab:nuscenes_overall_ablation} reports the effects of progressively adding PC-JEPA and $R^3$ to the MomAD baseline. PC-JEPA consistently reduces both long-horizon displacement error and collision rate, showing that candidate-conditioned future prediction benefits planning beyond the original temporal planner. Adding $R^3$ further improves all reported horizons, supporting the complementary contributions of relational future reasoning and residual reranking.

\noindent\textbf{Detailed PC-JEPA Ablation on NAVSIM v2.}
Table~\ref{tab:pcjepa_v2_detailed} reports component scores
for the core variants in Table~\ref{tab:pcjd_design}, with $R^3$
fixed. Baseline and Full match the reference and full
configurations in Table~\ref{tab:overall_planning_jepa}.
Shared Future, Single-Step Prediction, and w/o C--F Interaction
achieve 89.1, 90.0, and 90.3 EPDMS, respectively, compared
with 91.6 for Full PC-JEPA. These correspond to decreases
of 2.5, 1.6, and 1.3 points, matching the main ablation.
The full configuration has the highest EPDMS, while Shared
Future achieves higher EC and Single-Step Prediction achieves
higher HC, showing tradeoffs among the component metrics.

\noindent\textbf{Detailed $R^3$ Ablation on NAVSIM v2.}
Table~\ref{tab:r3_v2_detailed} expands the variants in
Table~\ref{tab:fcts_design}, with PC-JEPA fixed. Disabling
$R^3$ recovers the PC-JEPA-only configuration in
Table~\ref{tab:overall_planning_jepa}, with 89.6 EPDMS,
compared with 91.6 for Full $R^3$. Plan-Only Features,
Simple Concat., and w/o Compatibility achieve 90.5, 90.7,
and 90.0 EPDMS, respectively. Score Replacement reaches
90.1 EPDMS, 1.5 points below the full residual design.
Full $R^3$ achieves higher LK, HC, and EC scores of
97.7, 97.9, and 85.0, compared with 97.1, 97.6, and 84.9
for Score Replacement. The full configuration attains the
highest EPDMS and the highest or joint-highest component scores.

\subsection{Additional Qualitative Results}
\label{app:Qualitative}
\noindent\textbf{Additional Planning Visualizations on NAVSIM.}
Figure~\ref{vis_planning_nav} extends the qualitative analysis
in the main paper with additional driving scenarios. Across
varying road geometries and traffic interactions, ActWorld-JEPA
highlights planning-relevant semantic cues and generates
trajectories adapted to the surrounding scene. The overlays and trajectories provide complementary views
of how the selected plans relate to the surrounding scene.

\noindent\textbf{Long-Horizon Planning Visualization on nuScenes.}
Figure~\ref{nus_vis_fl} presents additional 6-second planning
results on the nuScenes validation set. Across diverse traffic
configurations, ActWorld-JEPA generates coherent trajectories
that adapt to road geometry and surrounding vehicles.
These examples illustrate the spatial relationship between
predicted trajectories, road structure, and nearby vehicles.
Together with the quantitative results, they provide
a qualitative view of long-horizon trajectory selection.

\end{document}